\documentclass[10pt,twocolumn,letterpaper]{article}

\usepackage{cvpr}
\usepackage{times}
\usepackage{epsfig}
\usepackage{graphicx}
\usepackage{amsmath}
\usepackage{amssymb}
\usepackage{subcaption}
\usepackage{algorithm}
\usepackage[noend]{algpseudocode}
\usepackage{booktabs}
\usepackage{multirow}
\usepackage{pifont}\usepackage[usenames, dvipsnames]{color}
\usepackage{bbm}

\usepackage[breaklinks=true,bookmarks=false]{hyperref}

\cvprfinalcopy 

\def\cvprPaperID{10437} 

\ifcvprfinal\fi
\begin{document}

\title{Seeing without Looking: Contextual Rescoring of Object Detections for AP Maximization}

\author{
    Lourenço V. Pato\textsuperscript{1}\\
    {\tt\small lourenco.pato@tecnico.ulisboa.pt}
  \and
    Renato Negrinho\textsuperscript{2}\\
    {\tt\small negrinho@cs.cmu.edu}
  \and
    Pedro M. Q. Aguiar\textsuperscript{1}\\
    {\tt\small aguiar@isr.ist.utl.pt}
  \and 
\textsuperscript{1}Institute for Systems and Robotics / IST, ULisboa
  \quad
    \textsuperscript{2}Carnegie Mellon University
}

\maketitle

\begin{abstract}
    The majority of current object detectors lack context: class predictions are made independently from other detections. We propose to incorporate context in object detection by post-processing the output of an arbitrary detector to rescore the confidences of its detections.
Rescoring is done by conditioning on contextual information from the entire set of detections: their confidences, predicted classes, and positions. 
    We show that AP can be improved by simply reassigning the detection confidence values such that true positives that survive longer (i.e., those with the correct class and large IoU) are scored higher than false positives or detections with small IoU.
    In this setting, we use a bidirectional RNN with attention for contextual rescoring and introduce a training target that uses the IoU with ground truth to maximize AP for the given set of detections.
The fact that our approach does not require access to visual features makes it computationally inexpensive and agnostic to the detection architecture.
In spite of this simplicity, our model consistently improves AP over strong pre-trained baselines (Cascade R-CNN and Faster R-CNN with several backbones),
particularly by reducing the confidence of duplicate detections (a learned form of non-maximum suppression) and removing out-of-context objects by conditioning on the confidences, classes, positions, and sizes of the co-occurrent detections. Code is available at {\small\url{https://github.com/LourencoVazPato/seeing-without-looking/}}
\end{abstract}

\section{Introduction}
The convolutional backbone of current object detectors processes the whole image to generate object proposals. However, these proposals are then classified independently, ignoring strong co-occurrence relationships between object classes.
By contrast, humans use a broad range of contextual cues to recognize objects~\cite{divvala:context}, such as class co-occurrence statistics and relative object locations and sizes.
This observation motivates our work, where we exploit contextual information from the whole set of detections to inform which detections to keep.

\begin{figure}[!t]
\begin{center}
   \includegraphics[height=5.8cm]{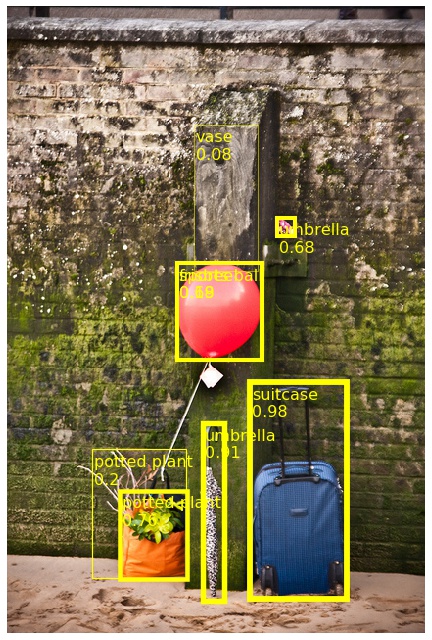}
   \includegraphics[height=5.8cm]{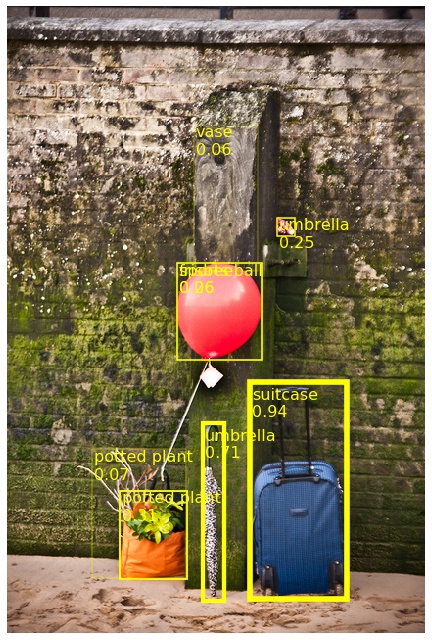}
\end{center}
   \caption{Detection confidences before (\textit{left}) and after (\textit{right}) contextual rescoring.
   High-confidence detections inform the topic of the image. 
   False positives have their confidences reduced (only suitcase and the umbrella are in the ground truth).
   The line thickness of a bounding box is proportional to its confidence.
   }
\label{fig:sample}
\end{figure}

\begin{figure*}[tbh]
\begin{center}
\includegraphics[width=\textwidth]{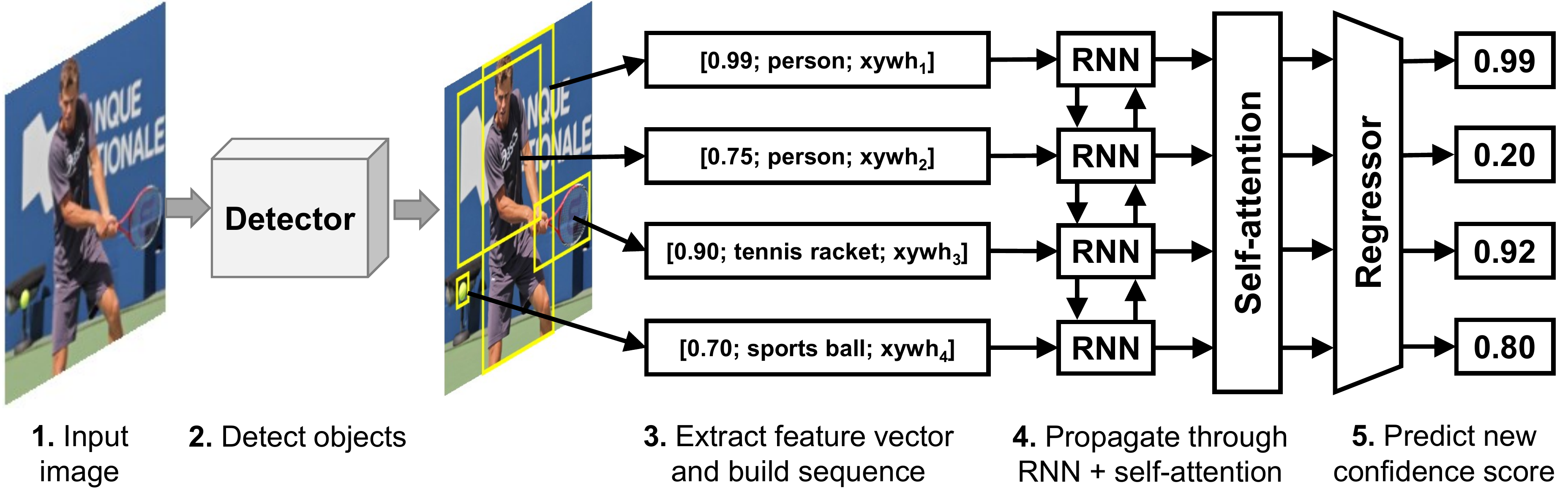}
\end{center}
   \caption{Overview of the contextual rescoring approach. \textbf{1-2.} A set of detections is collected by an object detector. \textbf{3.} A feature vector is extracted for each detection (by concatenating its confidence, predicted class, and coordinates). \textbf{4.} Detections are processed by an RNN with self-attention. \textbf{5.} A regressor predicts a new confidence for each detection. 
}
\label{fig:short}
\end{figure*}

Through an error analysis, we observe that current object detectors make errors that can be mitigated by the use of context. 
Errors can be ascribed to two types of problems: non-maximum suppression failing to remove duplicate detections (Figure~\ref{fig:duplicates_errors}); and local methods making insufficient use of context, e.g., when the object is visually similar to some class but the its context makes it unlikely (Figure~\ref{fig:out_of_context_errors}).

We first study how to improve AP by rescoring detections while keeping the same location and class (Section~\ref{optimal-rescoring}).
The insight is that detections with higher IoU count as true positives for more IoU thresholds and therefore should be scored higher. 
These scores are induced with the knowledge of the ground truth labels and lead to improvements of up to 15 AP on MS~COCO~\texttt{val2017} for detections produced by high-performance two-stage detectors (see Table~\ref{tab:target}).
Given a fixed matching between predicted and ground truth detections, to maximize AP, it is optimal to assign score equal to the IoU with the ground truth to each matched predicted detection.

We propose a model to rescore detections of a previous detector using context from all detections in the image (see Figure~\ref{fig:short}).
Each detection is represented by a feature vector with original confidence, predicted class, and bounding box coordinates. 
While the baseline detectors use only visual information, our model exploits non-visual high-level context, such as class co-occurrences, and object positions and sizes.
We use recurrent neural networks (RNNs) with self-attention to induce the contextual representation.
We train with a loss that pushes the model towards producing scores that maximize AP for the set of detections being rescored. 
Our approach is widely applicable as it does not use visual or other detector-specific features.

Results on MS~COCO~2017~\cite{COCO-dataset} (see Table~\ref{tab:rescored-results}) show that the proposed model improves AP by 0.5 to 1 across strong region-based baseline detectors (Faster R-CNN \cite{Ren:FasterRCNN} and Cascade R-CNN \cite{Cai:CascadeRCNN}) and different backbone networks (ResNet-101 and ResNet-50 \cite{He:ResNet}). Although the improvements may seem modest, we consider very strong baselines and obtain consistent improvements across them.
An analysis of the rescored detections (Section~\ref{sec:results}) shows that the model decreases the confidence for out-of-context and duplicate detections, while maintaining it for correct detections. 
Figure~\ref{fig:sample} illustrates this: false positives (sports ball, potted plant and umbrella) have their confidences reduced, while keeping high confidences for true positives (suitcase and umbrella).
We present additional examples picked systematically, i.e., those with the largest overall confidence changes according to the cosine distance (see Appendix~\ref{app:rescored-samples}).

We identify the following contributions of this work:
\begin{itemize}
\setlength{\itemsep}{0pt plus 1pt}
\item A rescoring algorithm to maximize AP given fixed sets of predicted and ground truth bounding boxes. 
We show that for detections produced by current two-stage object detectors, there is an improvement of approximately 15 AP.

    \item A contextual rescoring approach that generates a new confidence for each detection by conditioning on the confidences, classes, and positions of all detections. 
    Our model uses RNNs with self-attention to generate a contextual representation for each detection and it is trained to regress the values for AP maximization (i.e., IoU of the bounding box with the ground truth). 
\end{itemize}

\section{Related work}

\paragraph{Two-stage detectors}
State-of-the-art object detectors \cite{girshick:rcnn, Girshick:FastRCNN, Ren:FasterRCNN, Cai:CascadeRCNN} rely on a two-stage approach: select image regions likely to contain objects (e.g., using fixed region proposal algorithms \cite{girshick:rcnn, Girshick:FastRCNN} or a region proposal network \cite{Ren:FasterRCNN}) and then classify each region independently.
These approaches do not use non-visual contextual information.

\paragraph{Object detection with context}
Existing methods include context either in post-processing (as a rescoring or refinement step)~\cite{Felzenszwalb:DPM, Chen:Context-Refinement, choi:hierarchical-context, Cinibis:Set-based, Yu:theRoleofContext, divvala:context, Arbel:Inner-Scene} or in the detection pipeline~\cite{Mottaghi:Context, Bell:ION, Liu:Structure-inference-net, Li:AttentiveContexts, Chen:SpatialMemory, Ren:SingleStage}.
Existing work has incorporated context through multiple approaches such as logistic regression~\cite{divvala:context}, deformable parts-based models~\cite{Felzenszwalb:DPM,Mottaghi:Context}, latent SVMs~\cite{Yu:theRoleofContext}, binary trees~\cite{choi:hierarchical-context}, graphical models~\cite{Liu:Structure-inference-net}, spatial recurrent neural networks~\cite{Chen:SpatialMemory, Ren:SingleStage, Bell:ION} and skip-layer connections~\cite{Bell:ION}. 
Relation Networks~\cite{RelationNetworks} introduces a ``Object Relation Module'' that is incorporated into Faster R-CNN to capture inter-object relations and suppress duplicates.
Other work captures context by using RNNs to process visual feature maps~\cite{Li:AttentiveContexts, Chen:SpatialMemory, Ren:SingleStage, Bell:ION}.
Recently, \cite{utility2019barnea} explored the utility of context by rescoring detections using non-visual context inferred from ground truths. 
They consider how to improve AP by rescoring and propose an heuristic rule based on the ratio of true and false positives. Their approach does not provide a rescoring model as they condition on ground truth information.
To the best of our knowledge, we are the first to use a deep learning model that conditions on non-visual features (confidence, predicted class, and bounding box location) to rescore predictions generated by an arbitrary detector. Furthermore, our model is trained with a loss for AP maximization (see Section~\ref{optimal-rescoring}), which is developed based on the insight that better localized detections should be scored higher.

\begin{figure}[tbh]
    \centering
    \includegraphics[height=5.2cm]{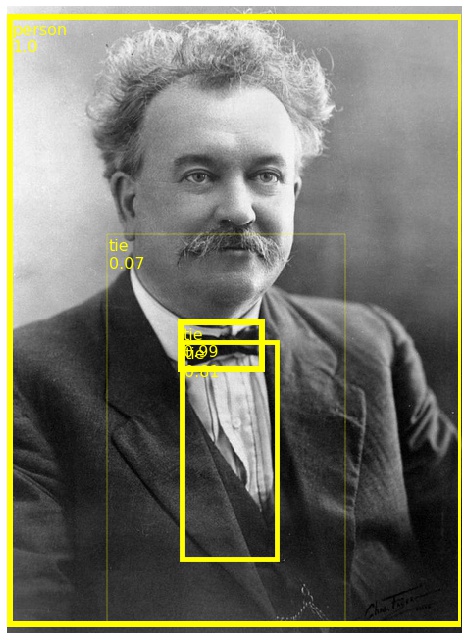}
    \includegraphics[height=5.2cm]{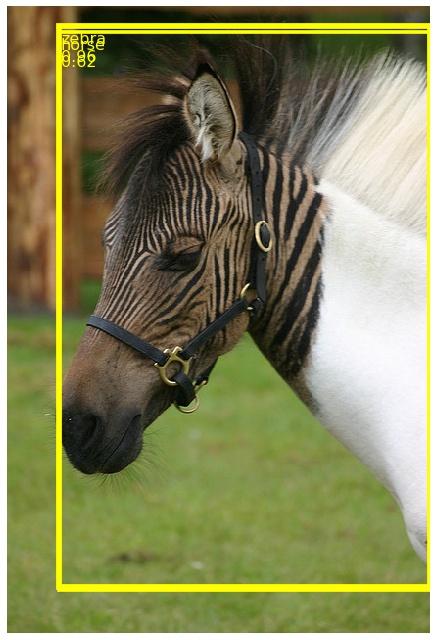}
    \caption{Duplicate detections illustrating failure cases of NMS. 
    \textit{Left:} Two high confidence detections of tie with low IoU. 
    \textit{Right:} Overlapping detections of horse and zebra.
} \label{fig:duplicates_errors}
\end{figure}

\paragraph{Non-maximum suppression}
NMS is a crucial component for removing duplicate detections. 
In addition to traditional NMS, Soft-NMS~\cite{Bodla:SoftNMS} reduces confidence proportionally to the IoU overlap, while learned NMS~\cite{hosang2015convnetNMS, hosang2017learning} learns the NMS rule from data. 
Both learned~NMS approaches use the same matching strategy used in evaluation and use a weighted logistic loss for rescoring (i.e., keep or remove a detection).
This loss does not encode preference for detections with better localization.
NMS approaches do not remove duplicate detections with different classes (Figure~\ref{fig:duplicates_errors} \textit{right}). 
By contrast, our approach conditions on all the predicted classes, confidences, and positions and therefore, our model can learn class, confidence and position-dependent suppression rules. 
Furthermore, we formulate a regression problem where the target is the IoU with ground truth such that better localized detections should be given a higher score.
In Section~\ref{optimal-rescoring}, we compare our rescoring approach (matching and targets) with learned~NMS approaches and show that there is large margin for improvement (Table~\ref{tab:target}).

\section{Error analysis}
\label{sec:error-analysis}

We analyze the errors made by two strong detectors.
For this analysis, we use the detections generated by MMDetection's~\cite{MMDetection} implementation of Faster R-CNN~\cite{Ren:FasterRCNN} and Cascade R-CNN~\cite{Cai:CascadeRCNN} with a ResNet-101~\cite{He:ResNet} backbone. The backbone is pre-trained for ImageNet~\cite{ILSVRC15} classification and fine-tuned for object detection on COCO \texttt{train2017}\footnote{For more information, please refer to the project's GitHub page \url{https://github.com/open-mmlab/mmdetection/}}.
Unless mentioned otherwise, all future analyses and examples will use results and examples from COCO \texttt{val2017} with Cascade R-CNN and a ResNet-101 backbone. 

\begin{figure}[tbh]
    \includegraphics[height=3.4cm]{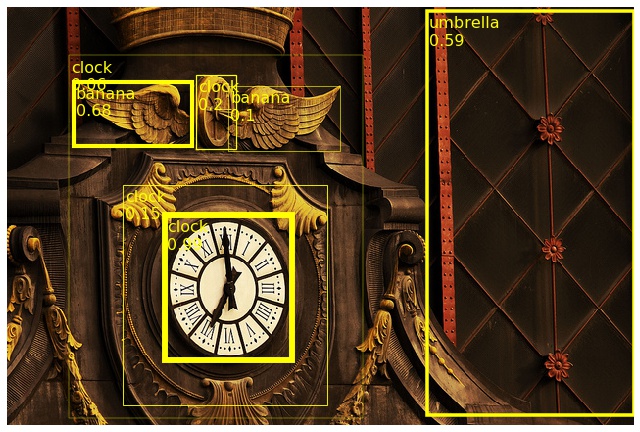}
    \includegraphics[height=3.4cm]{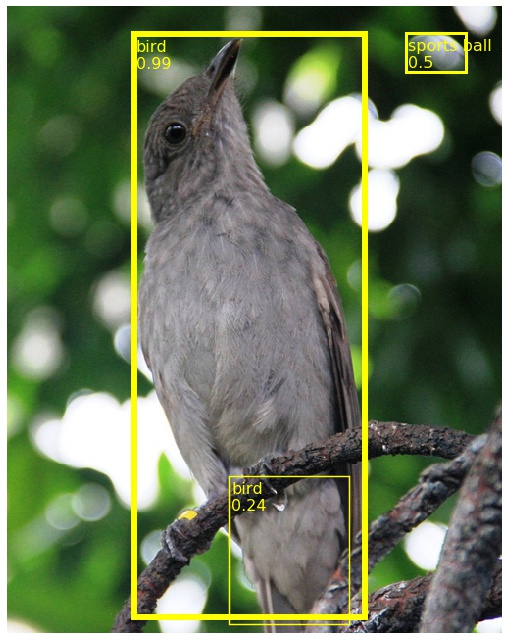}
    \caption{Failure cases of local non-contextual detection. 
    \textit{Left:} Banana and umbrella detected in a clock.
    \textit{Right:} Sports ball detected in the tree background.}
    \label{fig:out_of_context_errors}
\end{figure}

\subsection{Detection errors}

\paragraph{Localization errors and duplicate detections}
Localization errors occur when the predicted box has the correct class but low IoU with its ground truth, or when multiple boxes are predicted for the same object (duplicate detections).
NMS removes detections whose confidence is lower than any other detection with the same object class and IoU above a threshold (typically $0.7$~\cite{Ren:FasterRCNN}).
Unfortunately, NMS fails to remove duplicate detections with low IoU or with different classes, e.g., in Figure~\ref{fig:duplicates_errors}, a man with two ties (\textit{left}) and overlapping detections of zebra and horse (\textit{right}).
A learned contextual NMS procedure should suppress these false positives as it is unlikely for a person to have two ties and for a horse and a zebra to overlap completely. 

\begin{figure}[tbh]
    \begin{center}
    \includegraphics[width=\linewidth]{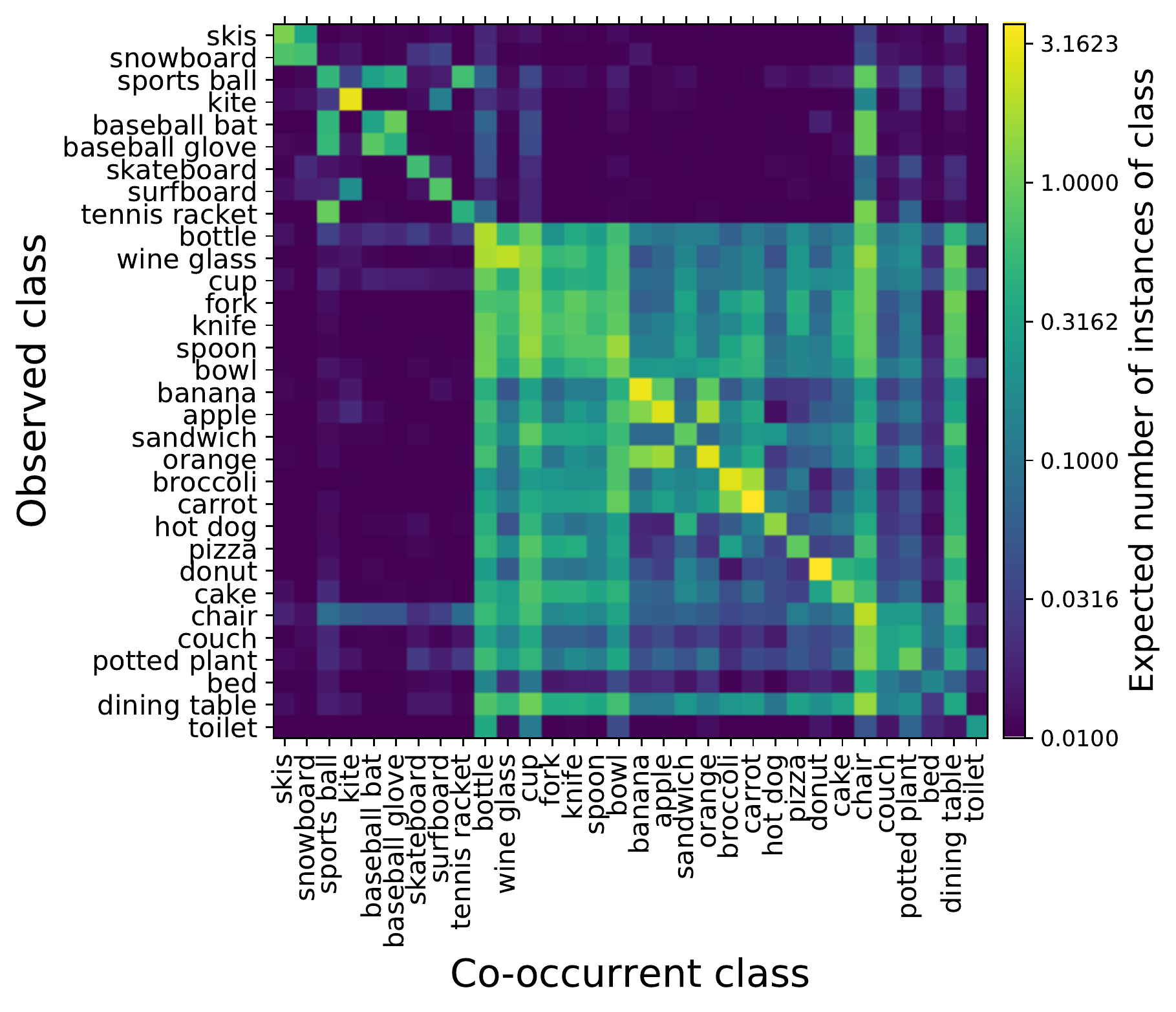}
    \end{center}
    \caption{Co-occurrences for a subset of classes in COCO \texttt{train2017}. Each cell represents the expected number of instances of the co-occurrent class in an image that has at least one instance from the observed class. 
    Related objects frequently co-occur: skis and snowboard; baseball bat, baseball glove and sports ball; cutlery.
    Rare co-occurrences are clear: sports objects and food rarely co-occur, bed and toilet appear with few other objects. There are strong diagonal co-occurrences: multiple classes frequently co-occur with themselves. Among these diagonal co-occurrences, toilet, bed and dining table are relatively weak.
    }        
\label{fig:cooccurrenceMatrix}
\end{figure}

\paragraph{Confusions with background and dissimilar class}
In Figure~\ref{fig:out_of_context_errors}, the detector finds unexpected objects such as an umbrella and a banana in a clock (\textit{left}), and a sports ball in a tree (\textit{right}).
A learned rescoring model should be able suppress these false positives due their low probability in their context, e.g., by capturing class co-occurrences.
Figure~\ref{fig:cooccurrenceMatrix} illustrates class co-occurrences for the ground truth objects in \texttt{val2017}. Each cell represents the expected number of instances of the co-occurrent class to be encountered in an image given that an instance from the observed class is present.
Using context, we can leverage these co-occurrences and decrease confidence for unexpected objects and increase it for detections that are likely correct. The figure with all class co-occurrences can be found in Appendix~\ref{app:cooccurrencematrix}.

\subsection{Statistical error analysis}
\label{ssec:statistical error analysis}

\begin{figure}[t]
    \centering
    \begin{subfigure}[b]{0.235\textwidth}
        \centering
        \includegraphics[width=\linewidth]{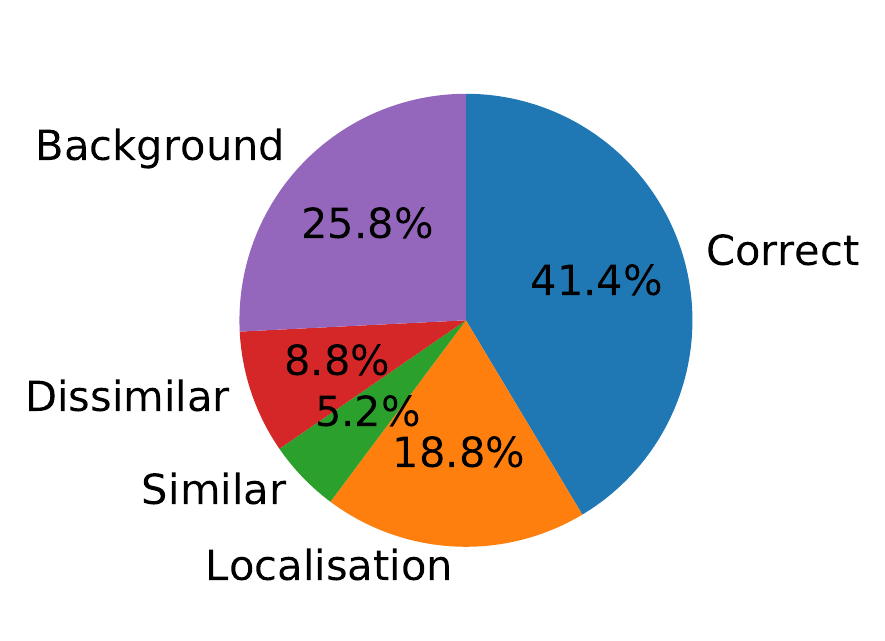}
        \caption{Faster R-CNN.}        
    \end{subfigure}
    \begin{subfigure}[b]{0.235\textwidth}
        \centering
        \includegraphics[width=\linewidth]{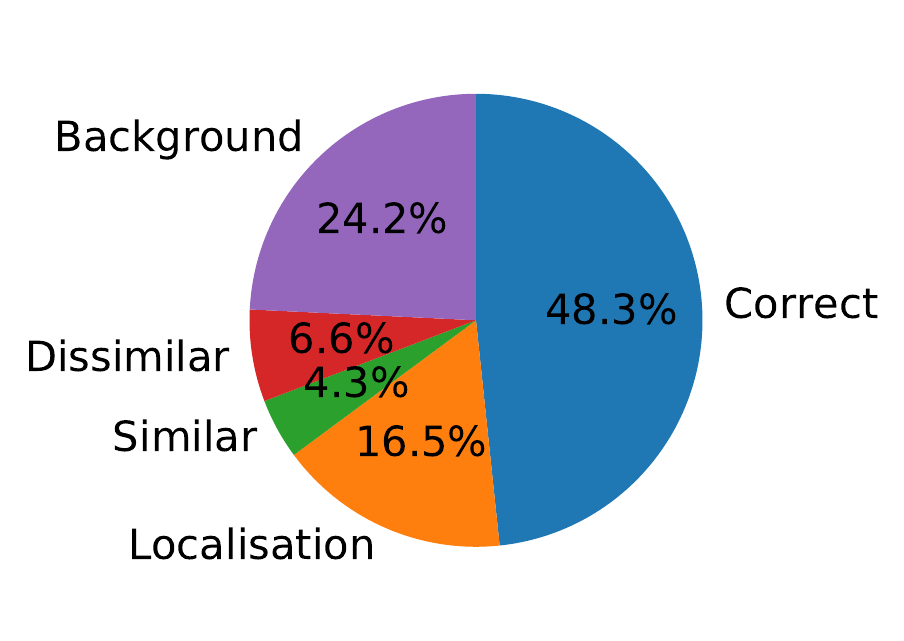}
        \caption{Cascade R-CNN.}        
    \end{subfigure}
    \caption{Confidence distribution of Faster R-CNN and Cascade R-CNN (ResNet-101 backbone) on \texttt{val2017}.}
    \label{fig:PieChart}
\end{figure}

Current object detectors place a significant amount of confidence on false positives (Figure~\ref{fig:PieChart}). 
We perform an analysis similar to~\cite{Hoiem:Diagnosing}, but because our rescoring approach does not change detections, only their scores, we change the metric to reflect the relative amount of confidence on each type of error.
Detections are split into five types:
\begin{itemize}
  \setlength{\itemsep}{0pt plus 1pt}
  \item \textbf{Correct:} correct class and location ($\text{IoU} \geq 0.5$).
  \item \textbf{Localization error:} correct class but wrong location ($0.1\leq\text{IoU}<0.5$); or correct location ($\text{IoU} \geq 0.5$), but ground truth already matched (duplicate detection). 
  \item \textbf{Confusion with similar class:} similar class (same COCO supercategory) and $\text{IoU}\geq 0.1$.
  \item \textbf{Confusion with dissimilar class:} dissimilar class (different COCO supercategory) and $\text{IoU}\geq 0.1$.
  \item \textbf{Confusion with background:} the remaining false positives ($\text{IoU}< 0.1$).
\end{itemize}
We iterated over detections by decreasing confidence and matched them with the ground truth with highest overlap, regardless of their class (by contrast, AP matches each class separately). 
In Figure~\ref{fig:PieChart}, we accumulate the total confidence placed on each type of detection (i.e., higher confidence detections have higher weight). Both Faster and Cascade R-CNN detectors place the majority of confidence on false positives. 
In Section~\ref{ssec:baselines} we compare the same distributions after rescoring and show that our rescoring model reduces the fraction of confidence placed on false positives (Figure~\ref{fig:piechart-confidences}) and increases AP (Table~\ref{tab:rescored-results}).

\section{Proposed approach: Contextual Rescoring}

We consider a simple post-processing strategy: keep the class and location of the predicted bounding boxes and change only their confidence. Detections can be removed by driving their confidence to zero. We show that given a set of detections and ground truth annotations, we can rescore detections such that AP is greatly improved (Table~\ref{tab:target}).

\subsection{Rescoring target}
\label{optimal-rescoring}

\paragraph{AP computation}
AP is computed for each class separately at various IoU thresholds ($0.5, 0.55, \dots, 0.95$). Increasing IoU thresholds reward better localization by requiring a detection to be closer to a ground truth to be considered true positive.
For computing AP, we first determine true and false positives by matching each detection with a ground truth.
COCO's matching strategy sorts detections by descending confidence order. 
Following this order, each detection is matched with the ground truth with the highest IoU if the following conditions are met: they have the same class, their IoU is greater or equal than the IoU threshold, and the ground truth was not yet matched. If no match is found, the detection is a false positive.

Then, the interpolated precision-recall curve is computed. 
Starting from the highest confidence detections, the curve $p(r)$ is traced by filling in the point that corresponds to the precision $p$ at the current recall $r$ for the running set of detections.
This curve is then made monotonically decreasing by re-assigning the precision at each recall level as the maximum precision at higher recalls:
\begin{equation}
  p_\text{interp}(r) = \operatorname*{\max}\limits_{\tilde{r} \geq r} p(\tilde{r}).
\end{equation}  
AP approximates the area under the interpolated precision-recall curve by averaging the interpolated precision at 101 equally spaced recall levels. For a given class $c$ and IoU threshold $t$, AP is given by
\begin{equation}
  \label{eq:pr-auc}
\text{AP}^{c}_t = \frac{1}{101} \sum_{r \in \{0, 0.01, \dots, 1\}} p_\text{interp}(r, c, t).
\end{equation}
The final metric for Average Precision is the average AP across the 80 object classes and at 10 different IoU levels,
\begin{equation}
  \label{eq:AP}
\text{AP} = \frac{1}{10} \sum_{t \in \{0.5, 0.55, \dots, 0.95\}} \frac{1}{80} \sum_{c \in \text{classes}} \text{AP}^{c}_t.
\end{equation}

\paragraph{Greedy maximization of AP}
Given a set of detections and ground truths, we aim to find the confidences that yield the maximum achievable AP.
To achieve this, we divide the maximization into \emph{two steps}: \emph{matching detections with ground truths} and \emph{selecting the optimal score for each detection}.
AP is a function of the ordering induced by the confidences but not their absolute values. Rescoring improves performance by reordering detections, assigning higher confidences to true positives than to false positives. 

\paragraph{Matching detections with ground truths}
Matching a detection with a ground truth is non-trivial because several detections can refer to the same ground truth. COCO's AP evaluation computes a different matching for each IoU threshold ($0.5, 0.55, \dots, 0.95$). For our rescoring approach, a single matching must be found.
A matching strategy that prioritizes detections by their confidence is penalized by AP when the highest confidence detection is not the best localized one. A high-confidence detection may be a true positive for lower IoU thresholds but become a false positive for higher thresholds. 
We propose an heuristic algorithm, Algorithm~\ref{alg:gt}, that prioritizes IoU with ground truth (i.e., better localization) over confidence. 
Starting from the highest IoU threshold and gradually reducing it (Line~\ref{alg:th}), the algorithm iterates over all ground truths (Line~\ref{alg:for}) and matches each ground truth with the detection with the highest overlap (Line~\ref{alg:argmax}) from the set of unmatched detections from the same class and with IoU above the threshold (Line~\ref{alg:conditions}). 
We denote the sets of already-matched predicted detections and ground truth detections as $\hat B(M) = \{\hat b \mid (\hat b, b^*) \in M\}$ and $B^*(M) = \{b^* \mid (\hat b, b^*) \in M \}$, respectively.

\begin{algorithm}[t]
    \caption{Greedy matching by ground truth overlap}
    \label{alg:gt}
    \begin{algorithmic}[1]
    \State{\textbf{Input:} Predicted detections $\hat B$, Ground truth $B^*$}
    \State{\textbf{Output:} Matching $M \subseteq \hat B \times B^*$}
    \State{$M \leftarrow \emptyset$}
    \For{$t \in \{0.95, 0.9, \dots, 0.5\}$} \label{alg:th} 
        \For{$b^* \in B^*$}  \label{alg:for}
            \If{$b^* \not \in B^*(M)$}
                \State $\hat{B}_{t, b^*} \leftarrow  \{\hat b \in \hat B \mid \text{class}(\hat b) = \text{class}(b^*), \hat b \not\in \hat{B}(M), \text{IoU}(\hat{b}, b^*) \geq t\}$ \label{alg:conditions}
                \If{$\hat{B}_{t, b^*} \neq \emptyset$}
                    \State{$b \leftarrow \operatorname*{\arg\max}\limits_{\hat b \in \hat{B}_{t, b^*}} \text{IoU}(\hat{b}, b^*)$} \label{alg:argmax}
                    \State{$M \leftarrow M \cup \{(b, b^*)\}$}
                \EndIf
            \EndIf
        \EndFor
    \EndFor
    \end{algorithmic}
\end{algorithm}

\begin{table}[t]
    \centering
    \begin{tabular}{@{}cccccc@{}}
    \toprule
    \multicolumn{1}{l}{Matching} & Target     & C-101           & C-50            & F-101           & F-50              \\ \midrule
    \multicolumn{1}{l}{}         & baseline   & 42.1           & 41.1           & 39.4           & 36.4             \\ \midrule
    \multirow{2}{*}{confidence}        & binary     & 47.8           & 46.9           & 44.8           & 42.9             \\
& IoU        & 55.4           & 54.5           & 52.8           & 51.0             \\ \midrule
    \multirow{2}{*}{localization} & binary     & 48.6           & 47.6           & 45.8           & 44.1             \\
& IoU        & \textbf{55.8}  & \textbf{54.9}  & \textbf{53.4}  & \textbf{51.7}    \\ \bottomrule
    \end{tabular}
    \caption{Average Precision for the target rescored values on \texttt{val2017}. 
\textbf{C:} Cascade R-CNN, \textbf{F:} Faster R-CNN, \textbf{101:} ResNet-101, \textbf{50:} ResNet-50. These rescoring results are computed from ground truths and predictions so they represent improvements achievable by an oracle. }
    \label{tab:target}
\end{table}

\paragraph{Optimal confidence values}
For a fixed matching, optimal rescoring orders detections such that those with higher IoUs have higher confidences. This ordering ensures that better localized detections have higher priority in AP's matching algorithm. 
Our proposed target confidence $y^*$ is the IoU with the matched ground truth for true positives and zero for false positives:
\begin{equation}
  \label{eq:target}
  y^*_{\hat b}=\begin{cases}
    \text{IoU} ( \hat{b}, b^* ) & \text{if $\hat{b} \in \hat{B}(M)$},\\
    0 & \text{otherwise},
  \end{cases}
\end{equation}
for $\hat b \in \hat G$ and $b^*$ is such that $(\hat b, b^*) \in M$.

\paragraph{Target AP}
Table~\ref{tab:target} compares the baseline AP obtained by Faster and Cascade R-CNN architectures (using ResNet-101 and ResNet-50 backbones) with the AP obtained if the detections are rescored using the proposed matching algorithms and target confidences.
Results are computed from the predictions and ground truths so they are only used to compute improved targets for training models.
Combinations in Table~\ref{tab:target} correspond to whether bounding boxes are greedily matched by the original confidence or IoU and whether the target confidence is binary (one if matched and zero otherwise) or its IoU with the ground truth.

Our matching strategy (Algorithm~\ref{alg:gt}) shows an improvement (ranging from $0.5$ to $1.5$) over a matching strategy that prioritizes confidence. 
Our target rescoring is around 8 AP better than the training target used by learned NMS approaches~\cite{hosang2015convnetNMS, hosang2017learning} (which use binary targets and confidence matching) and shows that large improvements (up to 15 AP) are possible just by rescoring detections.
In the following section, we train a rescoring model that uses contextual information to predict these target confidences.

\subsection{Model architecture}
We incorporate context to rescore detections produced by an earlier object detector (see Figure~\ref{fig:short}). The set of detections is mapped to a \textit{sequence} of features $\mathbf{x} \in \mathbb{R}^{L \times N}$ that is fed to our model that computes the rescored confidences $\mathbf{\hat y} \in \mathbb R^L$. Each rescored confidence in $\mathbf{\hat y}_i$ is generated by conditioning on $\mathbf x$ (i.e., the whole set of detections).

\paragraph{Feature extraction} 
A feature vector containing the original predicted confidence, class and location, is extracted for each detection in the image (see Equation~\ref{eq:features}). Together, they form a contextual representation for the set of detections.
For MS~COCO, the extracted feature vector is a 85-dimensional ($N = 85$) for detection $i$ is given by
\begin{equation}
    \mathbf{x}_i = [\mathit{score}_i] \oplus [\text{one\_hot} \left( \mathit{class}_i \right)] \oplus \left[ \frac{x_i}{W}, \frac{y_i}{H}, \frac{w_i}{W}, \frac{h_i}{H} \right],
    \label{eq:features}
\end{equation}
where $\oplus$ denotes vector concatenation, $x_i,~y_i$ are the coordinates of the top left corner of the detection bounding box, $w_i,~h_i$ are its width and height, and $W,~H$ are the width and height of the image. Features $\mathit{score}_i$ and $\mathit{class}_i$ are the detection confidence score and object class. Function \text{one\_hot} creates a one-hot vector encoding for the object class.
Detections are grouped by image and mapped to a sequence by sorting them by decreasing confidence. Sequences are padded to length 100 (the maximum number of detections often outputted by a detector). 

\paragraph{Recurrent neural networks}
The proposed model uses a bidirectional stacked GRU~\cite{cho2014GRU} to compute two hidden states $\overrightarrow{\mathbf{h}_t}$ and $\overleftarrow{\mathbf{h}_t}$ of size $n_h$, corresponding to the forward and backward sequences, that are concatenated to produce the state vector $\mathbf{h}_t$ of size $2 n_h$. We stack $n_r$ GRU layers.
The bidirectional model encodes each detection as a function of past and future detections in the sequence. 

\paragraph{Self-attention}
We use self-attention~\cite{Attention:is-all-you-need} to handle long range dependencies between detections which are difficult to capture solely with RNNs. 
For each element $i$, self-attention summarizes the whole sequence into a context vector $\mathbf{c}_i$, given by the average of all the hidden vectors in the sequence, weighted by an alignment score:
\begin{equation}
    \label{eq:context}
    \mathbf{c}_i = \sum_{j=1}^{L} \alpha_{ij} \mathbf{h}_j,
\end{equation}
where $L$ is the length of the sequence length before padding, $\mathbf{h}_j$ is the hidden vector of element $j$, and $\alpha_{ij}$ measures the alignment between $i$ and $j$.
The weights $\alpha_{ij}$ are computed by a softmax over the alignment scores:
\begin{equation}
    \label{eq:attention}
    \alpha_{ij} = \frac{\exp(\text{score}(\mathbf{h}_i, \mathbf{h}_j))}{\sum\limits_{k=1}^L \exp(\text{score}(\mathbf{h}_i, \mathbf{h}_k))},
\end{equation}
where $\text{score}(\mathbf{h}_i, \mathbf{h}_j)$ is a scoring function that measures the alignment between $\mathbf{h}_i$ and $\mathbf{h}_j$. We use the scaled dot-product \cite{Attention:is-all-you-need} function as a measure of alignment:
\begin{equation}
    \text{score}(\mathbf{h}_i, \mathbf{h}_j) = \frac{\mathbf{h}_{i}^\top \mathbf{h}_j}{\sqrt{L}}.
\end{equation}

\paragraph{Regressor}
Our model uses a multi-layer perceptron (MLP) to predict a value for the rescored confidence for each detection. The regressor input is the concatenation of the GRU's hidden vector $\mathbf{h}$ and the self-attention's context vector $\mathbf{c}$. Our proposed architecture consists of a linear layer of size $4n_h \times 80$ with ReLU activation, followed by a linear layer of size $80 \times 1$ with a sigmoid activation layer to produce an score between 0 and 1.

\paragraph{Loss function}
We formulate rescoring as regression for the target motivated by AP maximization (Section~\ref{optimal-rescoring}). We use squared error:
\begin{equation}
    \mathcal{L}(\mathbf{y}, \mathbf{y}^*) = \sum^{L}_{i=1} \left( \mathbf{y}_i - \mathbf{y}^*_i \right) ^2 ,
\end{equation}
where $\mathbf{y}$ are the rescored confidences, $\mathbf{y}^*$ is the target sequence computed by Algorithm~\ref{alg:gt} and Equation~\ref{eq:target}.

\begin{table*}[t]
    \centering
    \resizebox{\textwidth}{!}{\begin{tabular}{c|c||cccccc||cccccc}
    \bottomrule
    \multirow{2}{*}{\begin{tabular}[c]{@{}c@{}}Base model\\ (backbone)\end{tabular}}     &\multirow{2}{*}{rescored} & \multicolumn{6}{c||}{\texttt{val2017} (5k)}                           & \multicolumn{6}{c}{\texttt{test-dev2017} (20k)}                      \\ 
                                                                                         &                           &  AP   & AP$^{50}$ & AP$^{75}$ & AP$^S$ & AP$^M$ & AP$^L$ &  AP  & AP$^{50}$ & AP$^{75}$ & AP$^S$ & AP$^M$ & AP$^L$ \\ \midrule
    \multirow{2}{*}{\begin{tabular}[c]{@{}c@{}}Faster R-CNN\\ (ResNet-50)\end{tabular}}  &                  & 36.4 & 58.4     & 39.1     & 21.6  & 40.1  & 46.6  & 36.7 & 58.8      & 39.6      & 21.6   & 39.8   & 44.9   \\
                                                                                         & \ding{51}                 & \textbf{37.4} & \textbf{60.0}     & \textbf{40.1}     & \textbf{21.8}  & \textbf{40.7}  & \textbf{48.7}  & \textbf{37.4} & \textbf{60.2}      & \textbf{40.3}      & \textbf{21.8}  & \textbf{40.4}   & \textbf{46.1}   \\ \hline
    \multirow{2}{*}{\begin{tabular}[c]{@{}c@{}}Faster R-CNN\\ (ResNet-101)\end{tabular}} &                  & 39.4 & 60.7     & 43.0     & 22.1  & 43.6  & 52.0  & 39.7 & 61.4      & 43.2      & 22.1   & 43.1   & 50.2   \\
                                                                                         & \ding{51}                 & \textbf{39.9} & \textbf{61.6}     & \textbf{43.5}     & \textbf{22.4}  & \textbf{43.8}  & \textbf{53.0}  & \textbf{40.1} & \textbf{62.2}      & \textbf{43.5}      & 22.1   & \textbf{43.4}   & \textbf{50.8}   \\ \hline
    \multirow{2}{*}{\begin{tabular}[c]{@{}c@{}}Cascade R-CNN\\ (ResNet-50)\end{tabular}} &                  & 41.1 & 59.3     & 44.8     & 22.6  & 44.5  & 54.8  & 41.5 & 60.0      & 45.2      & 23.3   & 44.0   & 53.1   \\
                                                                                         & \ding{51}                 & \textbf{41.8} & \textbf{60.2}     & \textbf{45.3}     & \textbf{23.1}  & \textbf{45.1}  & \textbf{56.0}  & \textbf{42.0} & \textbf{60.7}      & \textbf{45.5}      & \textbf{23.5}   & \textbf{44.7}   & \textbf{54.2}   \\ \hline
    \multirow{2}{*}{\begin{tabular}[c]{@{}c@{}}Cascade R-CNN\\ (ResNet-101)\end{tabular}}&                  & 42.1 & 60.3     & 45.9     & 23.2  & 46.0  & 56.3  & 42.4 & 61.2      & 46.2      & 23.7   & 45.5   & 54.1   \\
                                                                                         & \ding{51}                 & \textbf{42.8} & \textbf{61.5}     & \textbf{46.5}     & \textbf{23.9}  & \textbf{46.7}  & \textbf{57.5}  & \textbf{42.9} & \textbf{62.1}      & \textbf{46.6}      & \textbf{23.9}   & \textbf{46.1}   & \textbf{55.3}   \\ \toprule                                                                                   
    \end{tabular}}
    \caption{Performance results before and after rescoring. AP$^S$, AP$^M$ and AP$^L$ refer to small, medium and large objects.}
    \label{tab:rescored-results}
\end{table*}

\begin{table}[tb]
 \centering
 \begin{tabular}{@{}cc|cc@{}}
 \toprule
 \multicolumn{2}{c|}{top positives} & \multicolumn{2}{c}{top negatives}      \\
 class        & $\Delta \text{AP}$  &  class              & $\Delta \text{AP}$     \\ \midrule
 toaster      & + 3.2               & wine glass          & - 0.4                  \\
 couch        & + 1.7               & person              & - 0.3                  \\
 hot dog      & + 1.6               & banana              & - 0.3                  \\
 frisbee      & + 1.4               & elephant            & - 0.3                  \\
 microwave    & + 1.4               & clock               & - 0.3                  \\
 baseball bat & + 1.4               & zebra               & - 0.2                  \\
 apple        & + 1.3               & tennis racket       & - 0.2                  \\
 sandwich     & + 1.2               & bicycle             & - 0.1                  \\
 pizza        & + 1.1               & bus                 & - 0.1                  \\
 cake         & + 1.1               & giraffe             & - 0.1                  \\ 
\bottomrule
 \end{tabular}
 \caption{Classes with highest changes in AP after rescoring.}
 \label{tab:AP-delta-acc}
\end{table}

\section{Experimental results}
\label{sec:results}

\subsection{Implementation details}

We ran existing detectors on MS~COCO~\cite{COCO-dataset} to generate detections for \texttt{train2017} (118k images) for training, \texttt{val2017} (5k images) for model selection, and \texttt{test-dev2017} (20k images) for evaluation.
As baseline detectors, we used MMDetection's \cite{MMDetection} implementations of Cascade R-CNN \cite{Cai:CascadeRCNN} and Faster R-CNN \cite{Ren:FasterRCNN} with ResNet-101 and ResNet-50 \cite{He:ResNet} backbones. 
We made our code available at \url{https://github.com/LourencoVazPato/seeing-without-looking} to easily train models on detections from arbitrary detectors.

\paragraph{Model hyperparameters}
The best hyperparameters found have hidden size $n_h=256$ and number of stacked GRUs $n_r=3$. We present model ablations in Appendix~\ref{app:ablations}.

\paragraph{Shuffling detections}
When a model is trained with input sequences ordered by descending confidence, it is biased into predicting the rescored confidences in the same decreasing order, yielding no changes to AP. We shuffle the input sequences during training with probability $0.75$.
As future work, it would be interesting to consider models that are invariant to the ordering of the bounding boxes.

\paragraph{Training} We use Adam with batch size $256$ and initial learning rate $0.003$.
When AP on the plateaus for more than $4$ epochs on \texttt{val2017} (i.e., the patience hyperparameter), the learning rate is multiplied by $0.2$ and the parameters are reverted to those of the best epoch. Training is stopped if validation AP does not improve for 20 consecutive epochs.

\subsection{Comparison with baselines}
\label{ssec:baselines}

Table~\ref{tab:rescored-results} compares performance before and after rescoring across different detectors. Rescored detections perform better, with consistent improvements ranging from 0.4 to 1 AP. Bigger objects achieve larger improvements ($\Delta\text{AP}^{L} > \Delta\text{AP}^{M} > \Delta\text{AP}^{S}$). 
Poorly localized detections have larger AP improvements ($\Delta\text{AP}^{50} > \Delta\text{AP}^{75}$).

\begin{figure}[tb]
    \centering
    \begin{subfigure}[b]{0.235\textwidth}
        \centering
        \includegraphics[width=\linewidth]{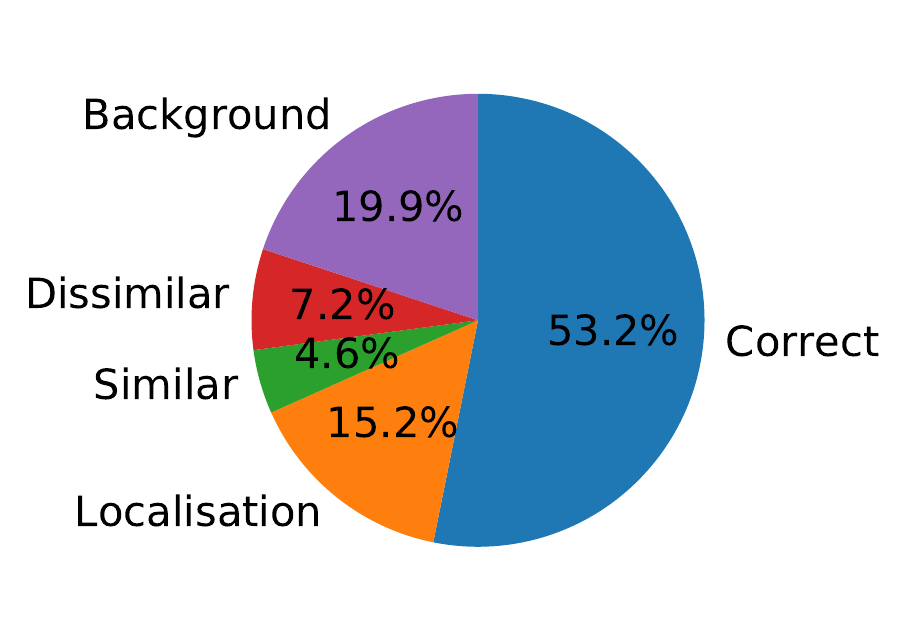}
        \caption{Faster R-CNN.}        
    \end{subfigure}
    \begin{subfigure}[b]{0.235\textwidth}
        \centering
        \includegraphics[width=\linewidth]{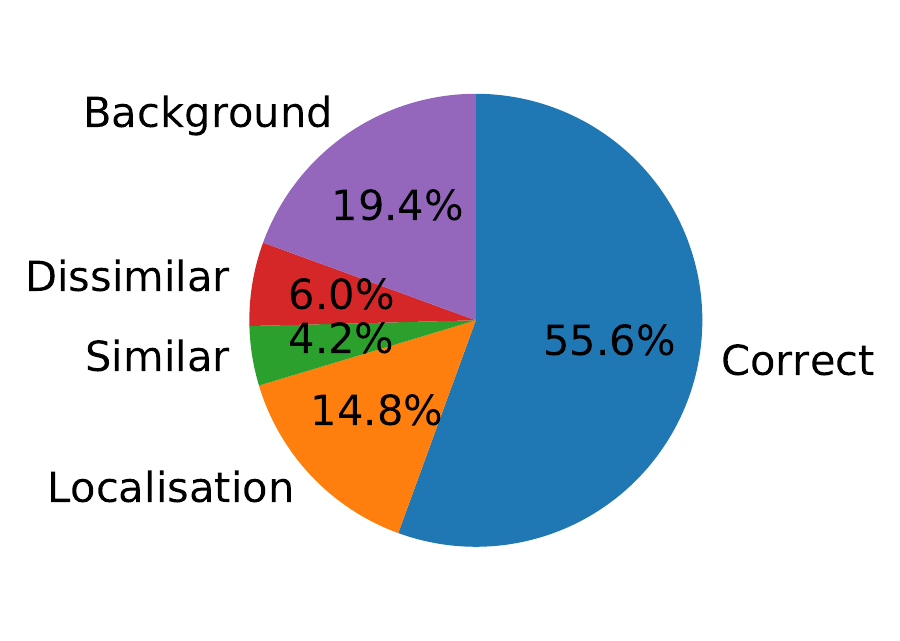}
        \caption{Cascade R-CNN.}        
    \end{subfigure}
    \caption{Accumulated confidence distribution on \texttt{val2017} after rescoring (compare to Figure~\ref{fig:PieChart}).}
    \label{fig:piechart-confidences}
\end{figure}

In Figure~\ref{fig:piechart-confidences}, we compare the total accumulated confidence for each error type, obtained by adding the confidence for all detections in \texttt{val2017} before and after rescoring (see Section~\ref{ssec:statistical error analysis}).
Correct detections have an increased share of the total confidence. Background and localization errors have a substantial reduction. 

\paragraph{Class AP}
Table~\ref{tab:AP-delta-acc} shows the classes with the largest changes in AP for Cascade R-CNN with ResNet-101 backbone. Other detectors can be found in Appendix~\ref{app:ablations}. Most classes show a significant and consistent AP increase.

\begin{table}[tb]
    \begin{center}
    \begin{tabular}{@{}c|cccc@{}}
    \toprule
        trained on         & \multicolumn{4}{c}{evaluated  on (\texttt{val2017})}                  \\
        \texttt{train2017} & F-50            & F-101           & C-50            & C-101          \\ \midrule
        F-50               & \textbf{+ 1.0}  & \textbf{+ 0.6}  & + 0.6           & + 0.5          \\
        F-101              & + 0.8           & + 0.5           & + 0.5           & + 0.5          \\
        C-50               & + 0.5           & + 0.1           & \textbf{+ 0.6}  & + 0.6          \\
        C-101              & + 0.5           & + 0.3           & + 0.5          & \textbf{+ 0.7} \\ \bottomrule
    \end{tabular}
    \caption{AP increase for models trained with different detectors (Faster R-CNN and Cascade R-CNN) and different backbones (ResNet-101 and ResNet-50). 
}
    \label{tab:generalisation}
    \end{center}
\end{table}

\paragraph{Generalization across architectures and backbones}
Different architectures have different error profiles. A rescoring model trained for one detector should hopefully generalize for other detectors. Table~\ref{tab:generalisation} compares the AP increase obtained by using a model trained on one detector and evaluated on a different one.
Although improvements are not as large when tested with different baselines, all models show consistent improvements.

\subsection{Ablations}

\paragraph{Training target}
Table~\ref{tab:trained-target} compares the AP achieved by our model when trained with a binary target and our proposed IoU target.
The difference in AP confirms that using the IoU with the ground truth better aligns with AP and produces higher improvements, as expected from Table~\ref{tab:target}.
\begin{table}[tbh]
    \centering
    \begin{tabular}{@{}ccccc@{}}
    \toprule
    target   & C-101 & C-50 & F-101 & F-50 \\ \midrule
    baseline & 42.1  & 41.1 & 39.4  & 36.4 \\ \midrule
    binary   & 42.5  & 41.6 & 39.6  & 37.3 \\
    IoU      & \textbf{42.8}  & \textbf{41.8} & \textbf{39.8}  & \textbf{37.4} \\ \bottomrule
    \end{tabular}
    \caption{Average Precision on COCO~\texttt{val2017} for binary and IoU training targets.} \label{tab:trained-target}
\end{table}

\paragraph{Feature importance}
Table~\ref{tab:feature-importance} explores feature importance by training the models with subsets of all the features. 
The most important feature is the original confidence, while the least important ones are the bounding box coordinates.
Not using the original confidence degrades AP by 2.2. 

\begin{table}[tbh]
    \centering
    \begin{tabular}{@{}ccccc@{}}
    \bottomrule
                     & conf. & class  & coord. & \texttt{val2017} AP \\ \midrule
    baseline         &            &           &             & 42.1               \\ \midrule
    all features     & \ding{51}  & \ding{51} & \ding{51}   & 42.8               \\ \midrule
    no coordinates   & \ding{51}  & \ding{51} &             & 42.4               \\
    no class         & \ding{51}  &           & \ding{51}   & 42.3               \\
    no confidence    &            & \ding{51} & \ding{51}   & 39.9               \\ \midrule
    just confidence  & \ding{51}  &           &             & 42.2               \\ \bottomrule
\end{tabular}
    \caption{Feature importance. The original confidence contributes the most to performance.}
    \label{tab:feature-importance}
\end{table}

\begin{figure}[!htb]
\begin{center}
   \includegraphics[height=5.2cm]{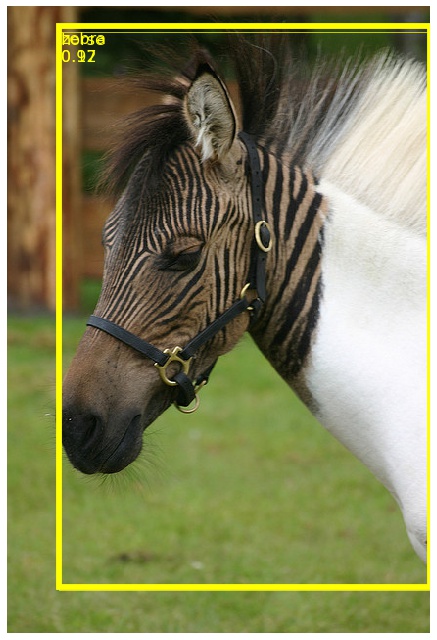}
   \includegraphics[height=5.2cm]{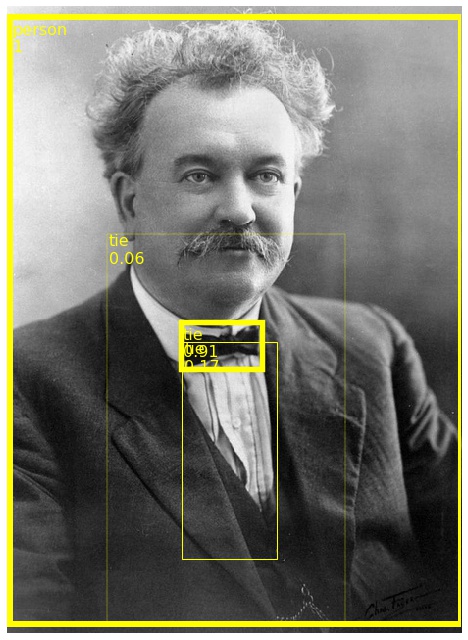}
\end{center}
   \caption{Detections after rescoring. Duplicate detections are suppressed (compare to Figure~\ref{fig:duplicates_errors}).}
\label{fig:rescored-duplicates}
\end{figure}

\begin{figure}[!htb]
\begin{center}
   \includegraphics[height=3.4cm]{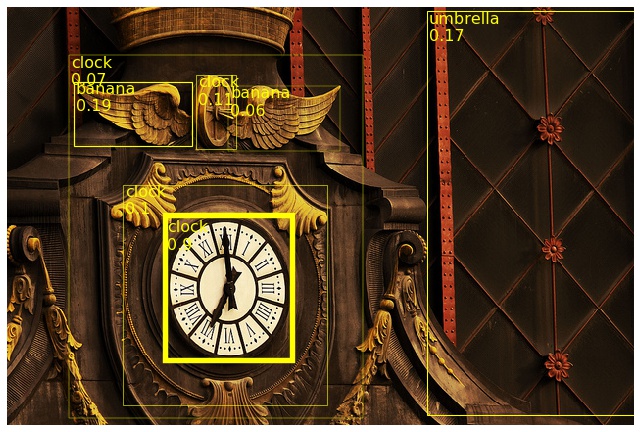}
   \includegraphics[height=3.4cm]{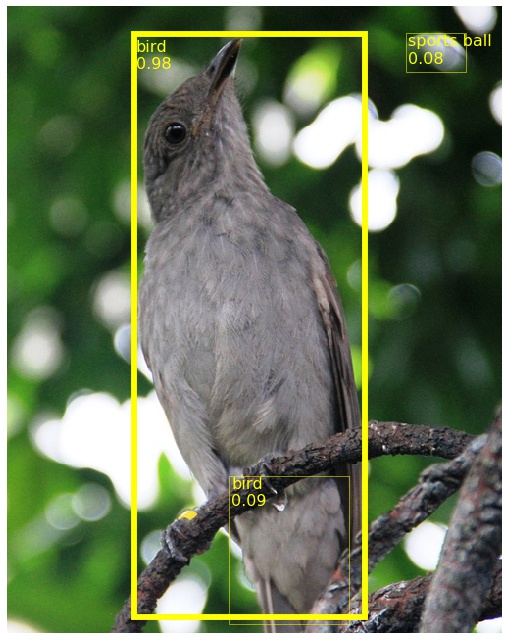}
\end{center}
   \caption{Detections after rescoring. False positives have been substantially suppressed (compare to Figure~\ref{fig:out_of_context_errors}).}
\label{fig:rescored-out-of-context}
\end{figure}

\section{Conclusions}
Current detectors make sub-optimal use of context, e.g., in a two-stage detector, each region is classified independently. Furthermore, NMS is an heuristic algorithm that fails to remove duplicates with low IoU or different classes. 
We observe that, to optimize AP, detections with better localization must be scored higher than poorly localized detections or false positives. 
Large increases in AP can be obtained solely by rescoring detections.
We train a contextual rescoring model, consisting of a bidirectional GRU with self-attention followed by a regressor, with this AP maximization target on MS~COCO. 
The experiments show that the model improves AP and reduces the total confidence placed on false positives across different baseline detectors. 
This model improves performance by 0.5 to 1 AP by exploiting solely non-visual context such as the confidences, classes, positions, and sizes of all detections in an image.

\paragraph{Acknowledgments} This work was partially supported by LARSyS - FCT Plurianual funding 2020-2023. We thank the anonymous reviewers for helpful comments.

{\small
\bibliographystyle{ieee_fullname}
\bibliography{mainbib}

\begin{thebibliography}{10}\itemsep=-1pt

\bibitem{Arbel:Inner-Scene}
Noa Arbel, Tamar Avraham, and Michael Lindenbaum.
\newblock Inner-scene similarities as a contextual cue for object detection.
\newblock {\em arXiv:1707.04406}, 2017.

\bibitem{utility2019barnea}
Ehud Barnea and Ohad Ben-Shahar.
\newblock Exploring the bounds of the utility of context for object detection.
\newblock In {\em CVPR}, 2019.

\bibitem{Bell:ION}
Sean Bell, C. Zitnick, Kavita Bala, and Ross Girshick.
\newblock Inside-outside net: Detecting objects in context with skip pooling
  and recurrent neural networks.
\newblock In {\em CVPR}, 2016.

\bibitem{Bodla:SoftNMS}
Navaneeth Bodla, Bharat Singh, Rama Chellappa, and Larry Davis.
\newblock Soft-nms--improving object detection with one line of code.
\newblock In {\em ICCV}, 2017.

\bibitem{Cai:CascadeRCNN}
Zhaowei Cai and Nuno Vasconcelos.
\newblock Cascade {R-CNN:} delving into high quality object detection.
\newblock In {\em CVPR}, 2018.

\bibitem{MMDetection}
Kai Chen, Jiaqi Wang, Jiangmiao Pang, Yuhang Cao, Yu Xiong, Xiaoxiao Li,
  Shuyang Sun, Wansen Feng, Ziwei Liu, Jiarui Xu, Zheng Zhang, Dazhi Cheng,
  Chenchen Zhu, Tianheng Cheng, Qijie Zhao, Buyu Li, Xin Lu, Rui Zhu, Yue Wu,
  Jifeng Dai, Jingdong Wang, Jianping Shi, Wanli Ouyang, Chen~Change Loy, and
  Dahua Lin.
\newblock {MMDetection}: Open mmlab detection toolbox and benchmark.
\newblock {\em arXiv:1906.07155}, 2019.

\bibitem{Chen:SpatialMemory}
Xinlei Chen and Abhinav Gupta.
\newblock Spatial memory for context reasoning in object detection.
\newblock In {\em ICCV}, 2017.

\bibitem{Chen:Context-Refinement}
Zhe Chen, Shaoli Huang, and Dacheng Tao.
\newblock Context refinement for object detection.
\newblock In {\em ECCV}, 2018.

\bibitem{cho2014GRU}
Kyunghyun Cho, Bart van Merri{\"e}nboer, Caglar Gulcehre, Fethi Bougares,
  Holger Schwenk, and Yoshua Bengio.
\newblock Learning phrase representations using {RNN} encoder-decoder for
  statistical machine translation.
\newblock {\em arXiv:1406.1078}, 2014.

\bibitem{choi:hierarchical-context}
Myung Choi, Joseph Lim, Antonio Torralba, and Alan Willsky.
\newblock Exploiting hierarchical context on a large database of object
  categories.
\newblock In {\em CVPR}, 2010.

\bibitem{Cinibis:Set-based}
Ramazan Cinbis and Stan Sclaroff.
\newblock Contextual object detection using set-based classification.
\newblock In {\em ECCV}, 2012.

\bibitem{divvala:context}
Santosh Divvala, Derek Hoiem, James Hays, Alexei Efros, and Martial Hebert.
\newblock An empirical study of context in object detection.
\newblock In {\em CVPR}, 2009.

\bibitem{Felzenszwalb:DPM}
Pedro Felzenszwalb, Ross Girshick, David McAllester, and Deva Ramanan.
\newblock Object detection with discriminatively trained part based models.
\newblock {\em TPAMI}, 2009.

\bibitem{Girshick:FastRCNN}
Ross Girshick.
\newblock Fast {R-CNN}.
\newblock In {\em ICCV}, 2015.

\bibitem{girshick:rcnn}
Ross Girshick, Jeff Donahue, Trevor Darrell, and Jitendra Malik.
\newblock Rich feature hierarchies for accurate object detection and semantic
  segmentation.
\newblock In {\em CVPR}, 2014.

\bibitem{He:ResNet}
Kaiming He, Xiangyu Zhang, Shaoqing Ren, and Jian Sun.
\newblock Deep residual learning for image recognition.
\newblock In {\em CVPR}, 2016.

\bibitem{Hoiem:Diagnosing}
Derek Hoiem, Yodsawalai Chodpathumwan, and Qieyun Dai.
\newblock Diagnosing error in object detectors.
\newblock In {\em ECCV}, 2012.

\bibitem{hosang2015convnetNMS}
Jan Hosang, Rodrigo Benenson, and Bernt Schiele.
\newblock A convnet for non-maximum suppression.
\newblock In {\em German Conference on Pattern Recognition}, 2016.

\bibitem{hosang2017learning}
Jan Hosang, Rodrigo Benenson, and Bernt Schiele.
\newblock Learning non-maximum suppression.
\newblock In {\em CVPR}, 2017.

\bibitem{RelationNetworks}
Han Hu, Jiayuan Gu, Zheng Zhang, Jifeng Dai, and Yichen Wei.
\newblock Relation networks for object detection.
\newblock In {\em CVPR}, 2018.

\bibitem{Li:AttentiveContexts}
Jianan Li, Yunchao Wei, Xiaodan Liang, Jian Dong, Tingfa Xu, Jiashi Feng, and
  Shuicheng Yan.
\newblock Attentive contexts for object detection.
\newblock In {\em IEEE Transactions on Multimedia}, 2016.

\bibitem{COCO-dataset}
Tsung-Yi Lin, Michael Maire, Serge Belongie, James Hays, Pietro Perona, Deva
  Ramanan, Piotr Doll{\'a}r, and C.~Lawrence Zitnick.
\newblock Microsoft {COCO}: Common objects in context.
\newblock In {\em ECCV}, 2014.

\bibitem{Liu:Structure-inference-net}
Yong Liu, Ruiping Wang, Shiguang Shan, and Xilin Chen.
\newblock Structure inference net: Object detection using scene-level context
  and instance-level relationships.
\newblock In {\em CVPR}, 2018.

\bibitem{Attention:alignment}
Minh{-}Thang Luong, Hieu Pham, and Christopher Manning.
\newblock Effective approaches to attention-based neural machine translation.
\newblock In {\em EMNLP}, 2015.

\bibitem{Mottaghi:Context}
Roozbeh Mottaghi, Xianjie Chen, Xiaobai Liu, Nam-Gyu Cho, Seong-Whan Lee, Sanja
  Fidler, Raquel Urtasun, and Alan Yuille.
\newblock The role of context for object detection and semantic segmentation in
  the wild.
\newblock In {\em CVPR}, 2014.

\bibitem{Ren:SingleStage}
Jimmy Ren, Xiaohao Chen, Jian-Bo Liu, Wenxiu Sun, Jiahao Pang, Qiong Yan,
  Yu-Wing Tai, and Li Xu.
\newblock Accurate single stage detector using recurrent rolling convolution.
\newblock In {\em CVPR}, 2017.

\bibitem{Ren:FasterRCNN}
Shaoqing Ren, Kaiming He, Ross Girshick, and Jian Sun.
\newblock Faster {R-CNN:} towards real-time object detection with region
  proposal networks.
\newblock In {\em NIPS}, 2015.

\bibitem{ILSVRC15}
Olga Russakovsky, Jia Deng, Hao Su, Jonathan Krause, Sanjeev Satheesh, Sean Ma,
  Zhiheng Huang, Andrej Karpathy, Aditya Khosla, Michael Bernstein, et~al.
\newblock {ImageNet} large scale visual recognition challenge.
\newblock {\em IJCV}, 2015.

\bibitem{Attention:is-all-you-need}
Ashish Vaswani, Noam Shazeer, Niki Parmar, Jakob Uszkoreit, Llion Jones, Aidan
  Gomez, Lukasz Kaiser, and Illia Polosukhin.
\newblock Attention is all you need.
\newblock In {\em NIPS}, 2017.

\bibitem{Yu:theRoleofContext}
Ruichi Yu, Xi Chen, Vlad Morariu, and Larry Davis.
\newblock The role of context selection in object detection.
\newblock {\em arXiv:1609.02948}, 2016.

\end{thebibliography}
}

\clearpage
\appendix

\section{Co-occurrence matrix}
\label{app:cooccurrencematrix}

\begin{figure*}[ptb]
    \centering
    \includegraphics[width=0.95\textwidth]{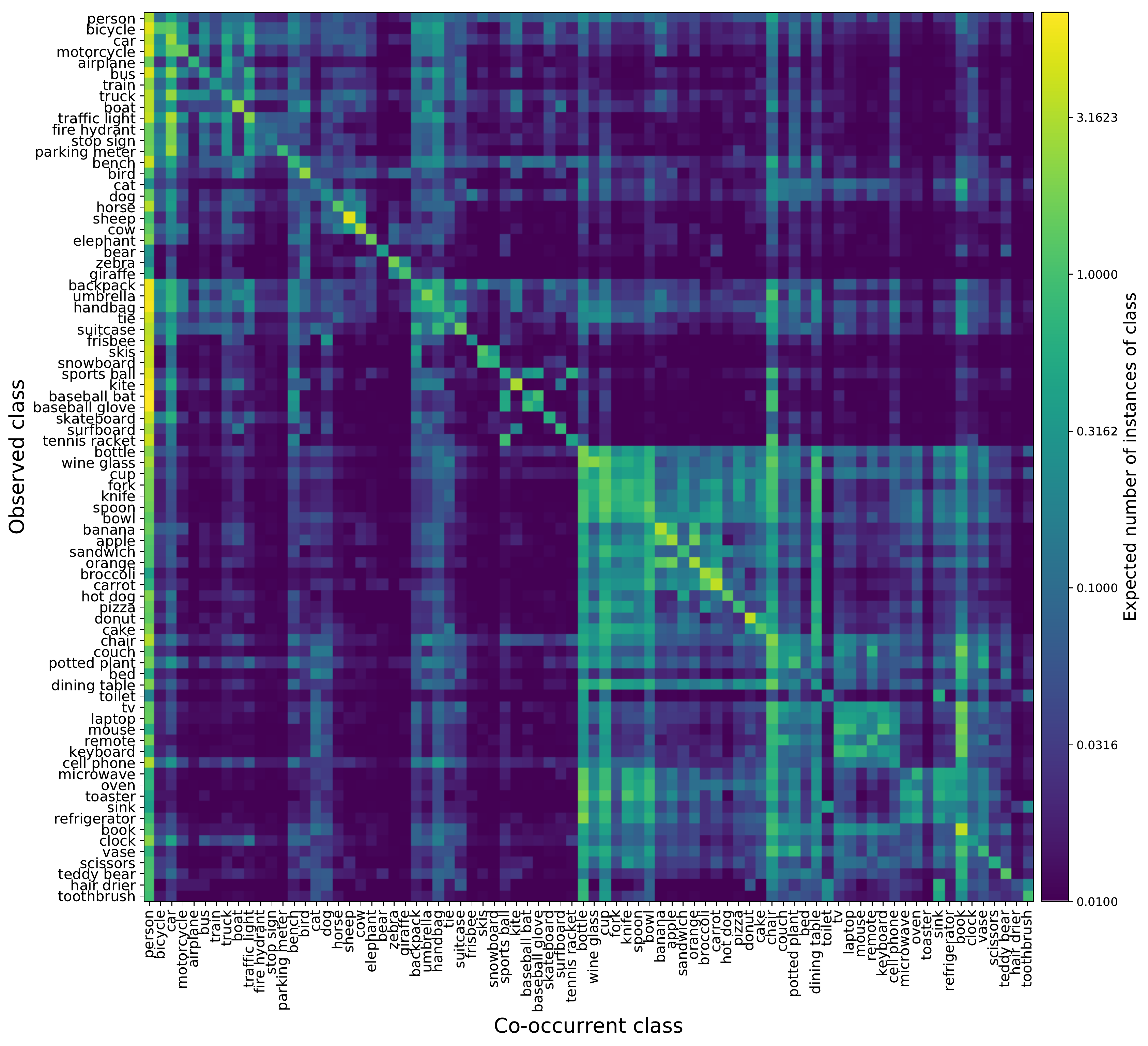}
    \caption{Co-occurrence matrix for COCO~\texttt{train2017} annotations. See Equation~\ref{eq:cooc-matrix} to see how these co-occurrences were computed.}
    \label{fig:cooccurrence-matrix-full}
\end{figure*}

Figure~\ref{fig:cooccurrence-matrix-full} illustrates all class co-occurrence statistics for the ground truth objects in COCO~\texttt{train2017}. 
Each entry represents the expected number of instances of the co-occurrent class given that there is at least one object from the observed class. 
If the co-occurrent and observed classes are the same, the entry represents how many co-occurrent instances of that class will be observed \emph{in addition} to the observed one.
Mathematically, for the set of classes $C = \{c_1, \ldots, c_k\}$, entry $(i, j) \in [k] \times [k]$ is computed as
\begin{equation}
 \label{eq:cooc-matrix}
 \frac{\sum _{q \in S_i} |c_j(G^*_q)| }{ |S_i| }  - \mathbbm{1}\{c _i = c_j \},
\end{equation}
where $S_i$ is the set of images containing at least one object of class $c_i$, i.e., $S_i = \{q \in [n] \mid |c_i(G_q^*)| \geq 1\}$.
Row $i$ iterates over observed classes, column $j$ iterates over co-occurrent classes, $G^*_q$ is the set of ground truth bounding boxes for image $q \in [n]$, where $n$ is the number of available images to compute the statistics (in this case, the number of images in \texttt{train2017}). $c(G^*_q) \subseteq G^*_q$ is the subset of bounding boxes in $G^*_q$ with class $c \in C$, and $\mathbbm{1}\{\cdot\}$ is the indicator function.

\section{Additional ablations and results}
\label{app:ablations}

\paragraph{Model comparison} 
Table~\ref{tab:modules} compares improvements obtained by using a bidirectional model and self-attention. The base model is an unidirectional RNN with $n_r=3$ stacked layers and a hidden state of size $n_h=256$ trained with shuffling instance with probability $0.75$. We compare the performance improvement of a bidirectional model and the addition of self-attention both with a GRU and a LSTM.
To compare to a model that does not use RNNs, we replace the RNN with a fully-connected layer (Linear(85,128) + ReLU) followed by self-attention (using ``general'' attention from \cite{Attention:alignment}) and the regressor (Linear(256,128) + ReLU + Linear(128,80) + ReLU + Linear(80,1) + Sigmoid). 

\begin{table}[tbh]
    \centering
    \begin{tabular}{@{}cccccc@{}}
    \bottomrule
    RNN    & attention & bidirectional & \# params & AP \\ \midrule
    baseline  &           &            &           & 42.1               \\ \midrule
    Linear    & \ding{51} &            & 0.1 M     & 42.6               \\ \midrule
    LSTM      &           &            & 1.4 M     & 42.6               \\
    LSTM      &           & \ding{51}  & 3.9 M     & 42.8               \\
    LSTM      & \ding{51} &            & 1.5 M     & 42.6               \\
    LSTM      & \ding{51} & \ding{51}  & 4.0 M     & 42.7               \\ \midrule
    GRU       &           &            & 1.1 M     & 42.6               \\
    GRU       &           & \ding{51}  & 2.9 M     & 42.8               \\
    GRU       & \ding{51} &            & 1.2 M     & 42.7               \\
    GRU       & \ding{51} & \ding{51}  & 3.0 M     & \textbf{42.8}      \\ \bottomrule
    \end{tabular}
    \caption{Ablation study of model components comparison. `Linear' replaces the RNN by a fully-connected layer.}
    \label{tab:modules}
\end{table}

The choice of LSTM or GRU has little impact on performance. GRU achieves higher performances with smaller models. 
Predictions made with an attention module or a bidirectional RNN conditions on the whole set of detections.
The results using a linear layer with self-attention demonstrate the attention mechanism's ability to capture context with fewer parameters.

\paragraph{Class AP improvements}
In Table~\ref{tab:class-AP-full} we aggregate the improvements on the per-class~AP for the tested baseline architectures on COCO~\texttt{test-dev2017}.
Our rescoring model produces consistent AP improvements for most classes, while few have a small decrease. 
The mean of each column is the improvement on the final AP metric for the model associated to that column. Faster R-CNN with a ResNet-50 backbone has the largest improvement

\begin{table*}[ptb]
        \begin{minipage}[t]{.49\textwidth}
            \centering
            \begin{tabular}{c|cccc|c}
            \textbf{class} & \textbf{C-101} & \textbf{C-50} & \textbf{F-101} & \textbf{F-50} & \textbf{Mean} \\ \hline
            couch & +1.7 & +1.4 & +1.9 & +2.5 & +1.88 \\  
            toaster & +3.2 & +0.9 & +1.2 & +1.5 & +1.70 \\  
            frisbee & +1.4 & +0.3 & +1.4 & +3.1 & +1.55 \\  
            cake & +1.1 & +1.4 & +1.3 & +1.7 & +1.38 \\  
            pizza & +1.1 & +1.0 & +1.9 & +1.3 & +1.32 \\  
            donut & +0.9 & +1.1 & +1.1 & +1.8 & +1.22 \\  
            sandwich & +1.2 & +1.2 & +1.0 & +1.1 & +1.13 \\  
            orange & +1.1 & +1.3 & +0.4 & +1.7 & +1.13 \\  
            toilet & +0.6 & +0.7 & +1.4 & +1.7 & +1.10 \\  
            bed & +0.4 & +0.3 & +1.7 & +2.0 & +1.10 \\  
            refrigerator & +0.6 & +0.8 & +1.1 & +1.9 & +1.10 \\  
            microwave & +1.4 & +1.0 & +0.7 & +0.9 & +1.00 \\  
            vase & +1.0 & +0.7 & +1.4 & +0.9 & +1.00 \\  
            hair drier & +0.1 & +0.2 & +2.9 & +0.7 & +0.98 \\  
            laptop & +0.6 & +1.0 & +1.0 & +0.8 & +0.85 \\  
            carrot & +1.0 & +1.0 & +0.5 & +0.9 & +0.85 \\  
            mouse & +0.8 & +0.8 & +0.6 & +1.2 & +0.85 \\  
            cow & +0.8 & +0.8 & +0.5 & +1.2 & +0.83 \\  
            surfboard & +0.6 & +0.9 & +0.5 & +1.3 & +0.83 \\  
            baseball glove & +0.8 & +0.5 & +0.9 & +1.1 & +0.83 \\
            snowboard & +0.4 & +0.7 & +0.6 & +1.6 & +0.83 \\
            cell phone & +0.8 & +0.6 & +0.9 & +0.9 & +0.80 \\
            dining table & +0.4 & +0.3 & +0.6 & +1.9 & +0.80 \\
            baseball bat & +1.4 & +0.6 & -0.2 & +1.3 & +0.78 \\
            cat & +0.2 & +0.3 & +0.8 & +1.7 & +0.75 \\
            fork & +0.6 & +0.8 & +0.3 & +1.3 & +0.75 \\
            backpack & +0.6 & +0.5 & +0.7 & +1.0 & +0.70 \\
            broccoli & +0.8 & +0.4 & +0.4 & +1.2 & +0.70 \\
            toothbrush & +0.6 & +1.2 & +0.4 & +0.6 & +0.70 \\
            bench & +0.5 & +0.6 & +0.8 & +0.8 & +0.68 \\
            suitcase & +0.4 & +0.7 & +0.3 & +1.2 & +0.65 \\
            oven & +0.3 & +1.1 & +0.3 & +0.8 & +0.63 \\
            remote & +0.7 & +0.8 & +0.3 & +0.6 & +0.60 \\
            kite & +0.5 & +0.3 & +0.4 & +1.1 & +0.58 \\
            apple & +1.3 & +0.5 & +0.6 & -0.2 & +0.55 \\
            knife & +0.6 & +0.7 & +0.3 & +0.6 & +0.55 \\
            teddy bear & +0.7 & +1.1 & -0.1 & +0.5 & +0.55 \\
            scissors & +0.3 & +1.1 & -0.3 & +1.0 & +0.53 \\
            skis & +0.4 & +0.6 & +0.1 & +0.9 & +0.50 \\
            hot dog & +1.6 & +0.3 & +0.3 & -0.2 & +0.50 \\
            truck & +0.3 & +0.2 & -0.1 & +1.4 & +0.45 \\
            \end{tabular}
        \end{minipage}
        \begin{minipage}[t]{.49\textwidth}
            \centering
            \begin{tabular}{c|cccc|c}
            \textbf{class} & \textbf{C-101} & \textbf{C-50} & \textbf{F-101} & \textbf{F-50} & \textbf{Mean} \\ \hline
            horse & +0.4 & +0.4 & +0.0 & +1.0 & +0.45 \\
            umbrella & +0.4 & +0.4 & +0.5 & +0.5 & +0.45 \\
            skateboard & +0.4 & +0.2 & +0.3 & +0.8 & +0.43 \\
            spoon & +0.7 & +0.2 & +0.4 & +0.4 & +0.43 \\
            chair & +0.4 & +0.4 & +0.3 & +0.5 & +0.40 \\
            parking meter & +0.4 & +0.8 & -0.6 & +1.0 & +0.40 \\
            train & +0.1 & +0.5 & +0.0 & +1.0 & +0.40 \\
            cup & +0.3 & +0.4 & +0.3 & +0.5 & +0.38 \\
            bear & +0.6 & +0.4 & +0.0 & +0.4 & +0.35 \\
            tv & +0.4 & +0.1 & +0.1 & +0.7 & +0.33 \\
            car & +0.3 & +0.1 & +0.3 & +0.5 & +0.30 \\
            bowl & +0.3 & +0.4 & +0.3 & +0.1 & +0.28 \\
            fire hydrant & +0.1 & +0.5 & +0.0 & +0.5 & +0.28 \\
            airplane & +0.1 & +0.3 & -0.1 & +0.6 & +0.23 \\
            dog & +0.8 & +0.2 & -0.1 & -0.1 & +0.20 \\
            sports ball & +0.3 & +0.3 & -0.1 & +0.2 & +0.18 \\
            sheep & +0.2 & +0.1 & +0.2 & +0.2 & +0.18 \\
            keyboard & +0.0 & +0.6 & -0.3 & +0.3 & +0.15 \\
            handbag & +0.1 & +0.2 & +0.1 & +0.0 & +0.10 \\
            bottle & +0.1 & +0.0 & +0.4 & -0.1 & +0.10 \\
            tie & +0.6 & +0.0 & -0.1 & -0.2 & +0.08 \\
            banana & -0.3 & +0.4 & -0.2 & +0.3 & +0.05 \\
            stop sign & +0.1 & +0.3 & -0.3 & +0.1 & +0.05 \\
            book & +0.0 & +0.0 & +0.1 & +0.0 & +0.03 \\
            potted plant & +0.1 & -0.1 & -0.4 & +0.5 & +0.03 \\
            bus & -0.1 & -0.1 & -0.2 & +0.5 & +0.03 \\
            boat & +0.0 & -0.2 & +0.1 & +0.1 & +0.00 \\
            bird & +0.1 & -0.3 & -0.1 & +0.3 & +0.00 \\
            motorcycle & +0.1 & -0.1 & -0.2 & +0.1 & -0.03 \\
            tennis racket & -0.2 & +0.1 & -0.5 & +0.4 & -0.05 \\
            traffic light & +0.0 & +0.0 & -0.3 & -0.1 & -0.10 \\
            zebra & -0.2 & -0.2 & -0.2 & +0.2 & -0.10 \\
            sink & +0.2 & +0.2 & -0.4 & -0.7 & -0.18 \\
            bicycle & -0.1 & -0.3 & -0.4 & +0.0 & -0.20 \\
            person & -0.3 & -0.3 & -0.2 & +0.0 & -0.20 \\
            wine glass & -0.4 & -0.2 & -0.2 & +0.0 & -0.20 \\
            giraffe & -0.1 & -0.4 & -0.7 & -0.1 & -0.33 \\
            elephant & -0.3 & -0.3 & -0.8 & -0.4 & -0.45 \\
            clock & -0.3 & -0.5 & -0.8 & -0.3 & -0.48 \\ \hline
Mean & +0.5 & +0.5 & +0.4 & +0.7 & +0.53 \\
            \end{tabular}
        \end{minipage}
    \caption{Per-class AP improvement on \texttt{test-dev2017}. \textbf{C:} Cascade R-CNN, \textbf{F:} Faster R-CNN, \textbf{101:} ResNet-101, \textbf{50:} ResNet-50.}
    \label{tab:class-AP-full}
\end{table*}

\section{Rescored examples}
\label{app:rescored-samples}

To systematically explore the results of rescoring, we compare, for each image, the vectors of confidences for the detections before and after rescoring. We sort images in decreasing order of the change in confidences, as measured by the cosine distance between the vectors of confidences before and after rescoring, i.e., for image $q \in [n]$, 
\begin{equation}
    d(v_q, v_q') = 1 - \frac{v_q ^T v_q'}{||v_q||_2||v_q'||_2}, 
\end{equation}
where $v_q, v'_q \in \mathbb{R}^{|\hat{G}_q|}$ are the vectors of confidences before and after rescoring, respectively, and $\hat G_q$ is the set of detections being rescored. This analysis uses the detections produced by Cascade R-CNN with a ResNet-101 backbone on \texttt{val2017}.

\begin{figure*}[tbh]
    \centering
    \begin{minipage}{.49\linewidth}
            \includegraphics[width=0.32\linewidth]{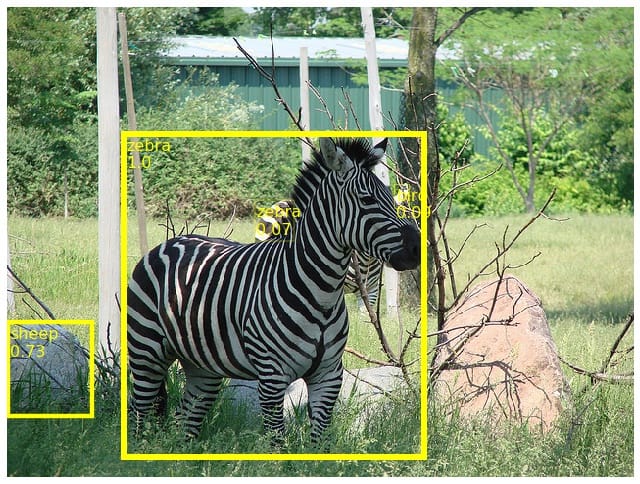}
            \includegraphics[width=0.32\linewidth]{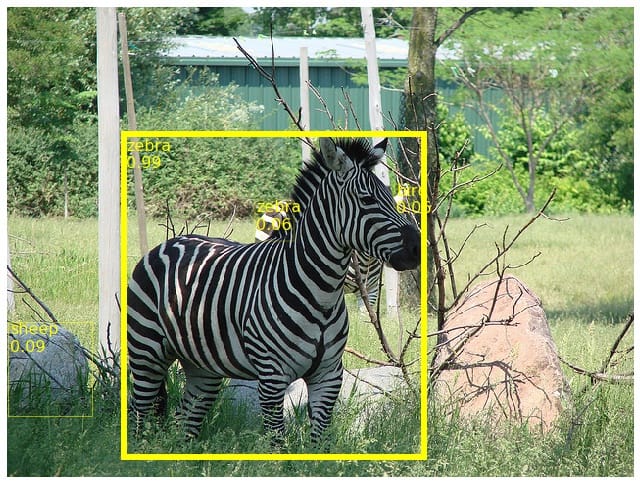}
            \includegraphics[width=0.32\linewidth]{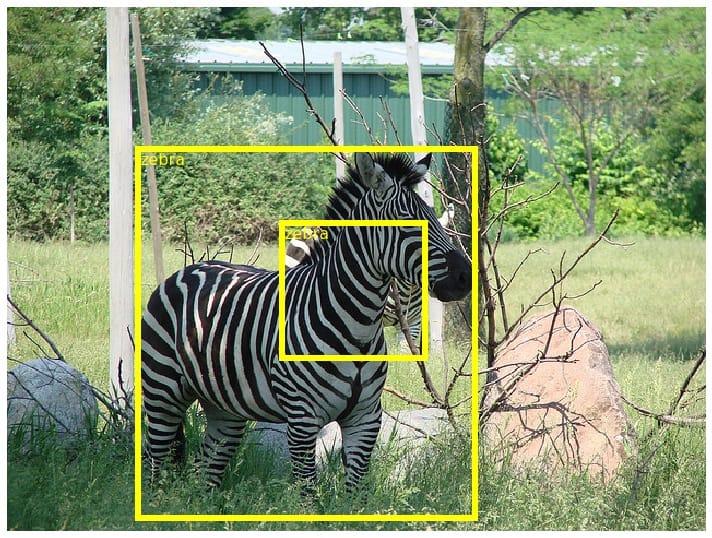}
            \includegraphics[width=0.32\linewidth]{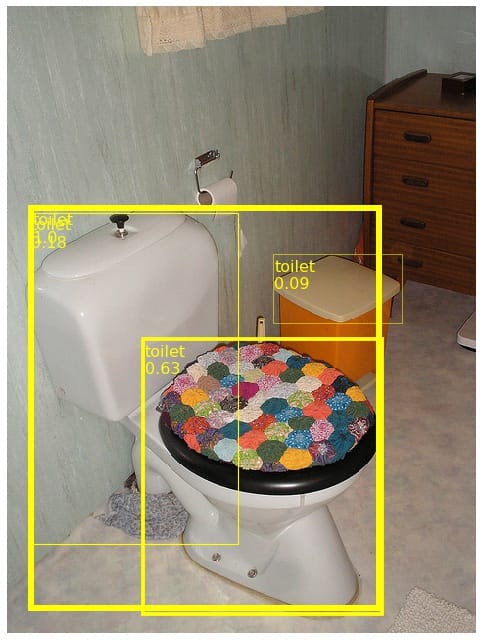}
            \includegraphics[width=0.32\linewidth]{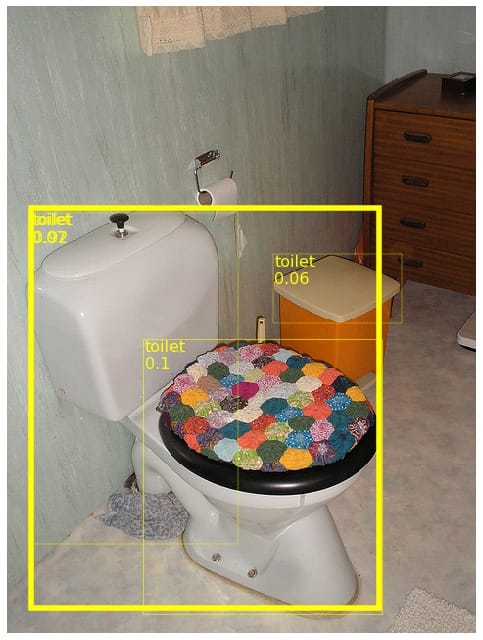}
            \includegraphics[width=0.32\linewidth]{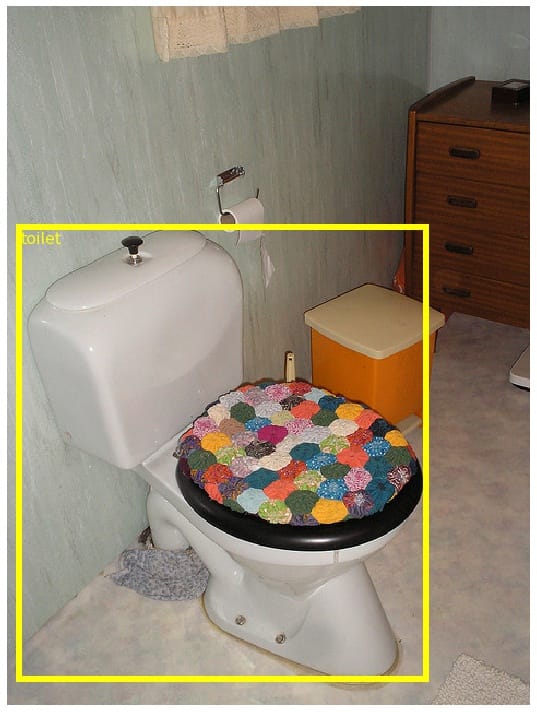}
            \includegraphics[width=0.32\linewidth]{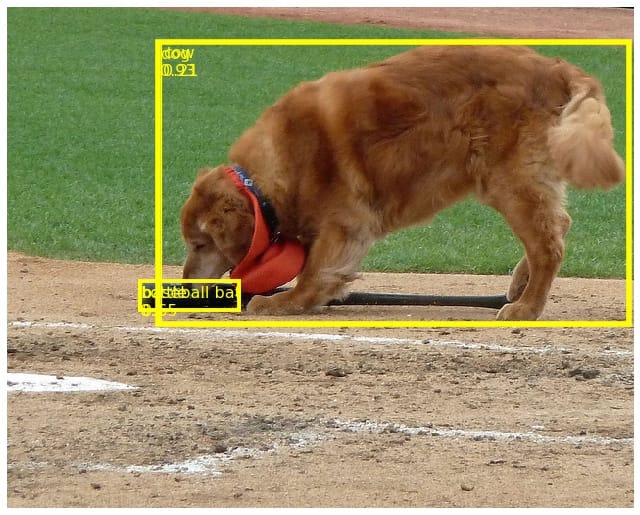}
            \includegraphics[width=0.32\linewidth]{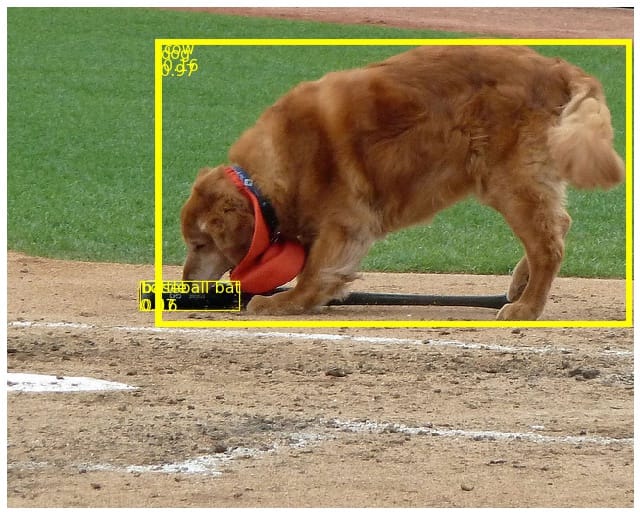}
            \includegraphics[width=0.32\linewidth]{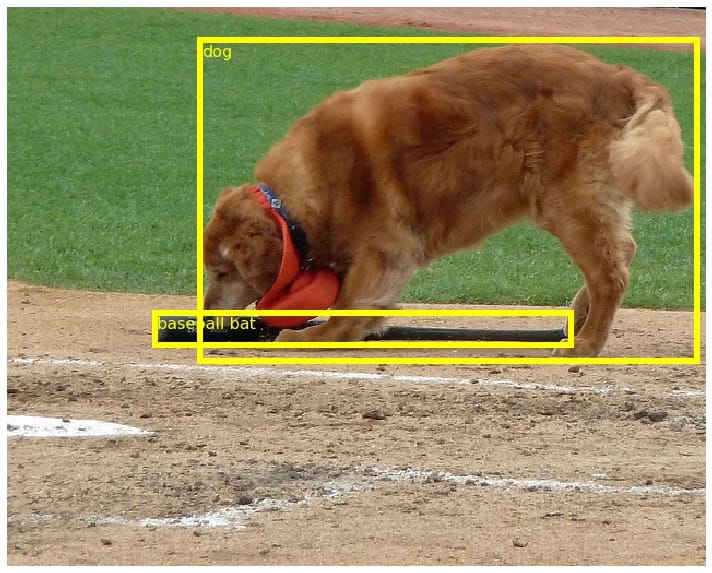}
            \includegraphics[width=0.32\linewidth]{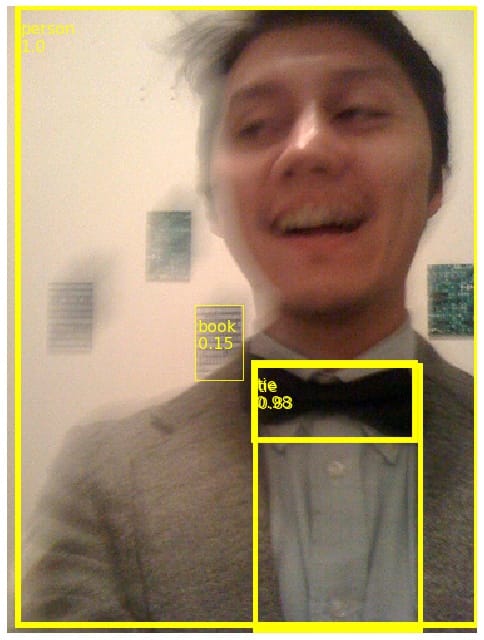}
            \includegraphics[width=0.32\linewidth]{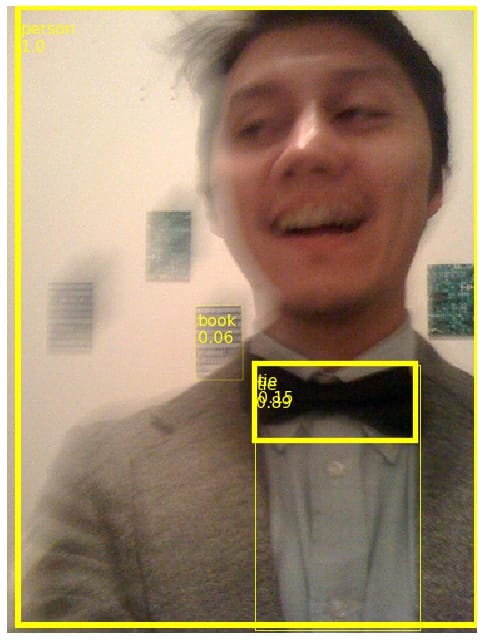}
            \includegraphics[width=0.32\linewidth]{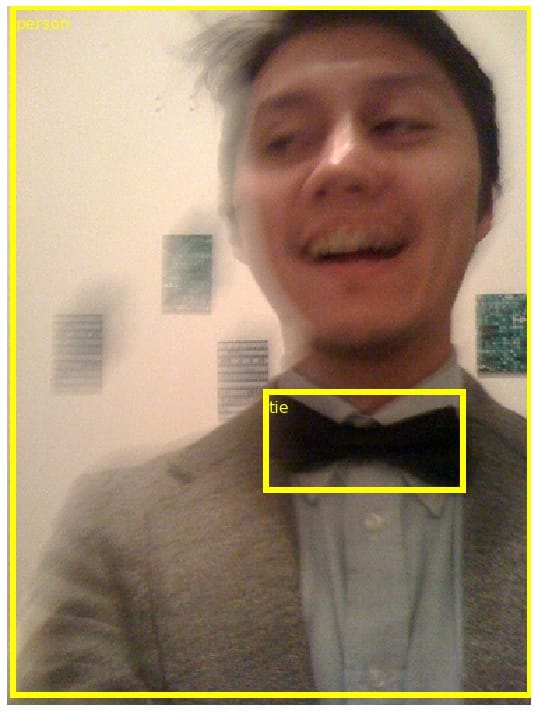}
            \includegraphics[width=0.32\linewidth]{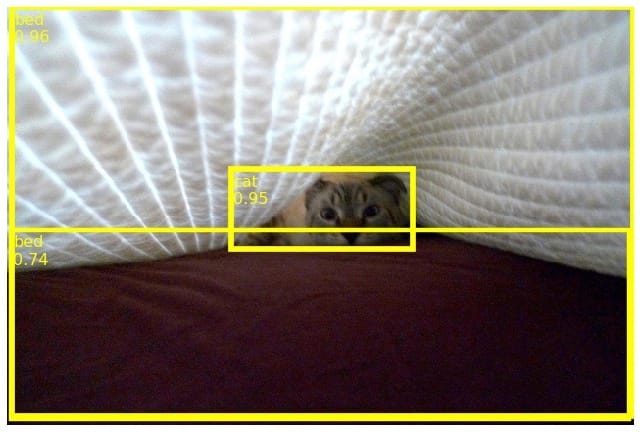}
            \includegraphics[width=0.32\linewidth]{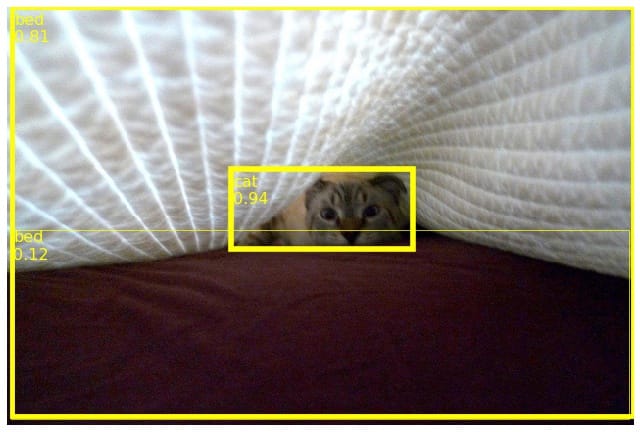}
            \includegraphics[width=0.32\linewidth]{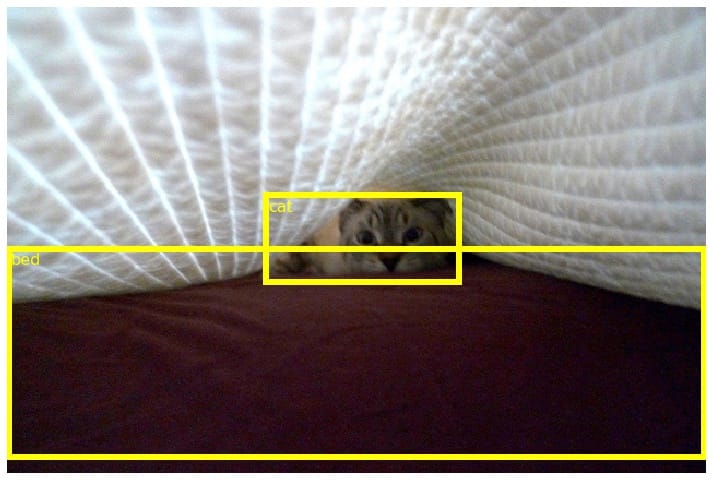}    
            \includegraphics[width=0.32\linewidth]{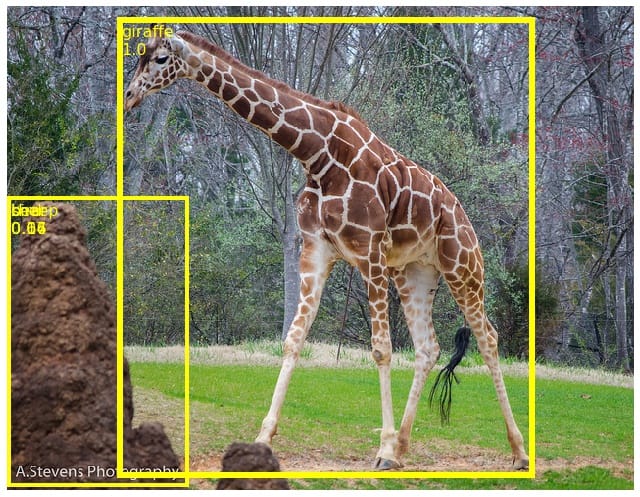}
            \includegraphics[width=0.32\linewidth]{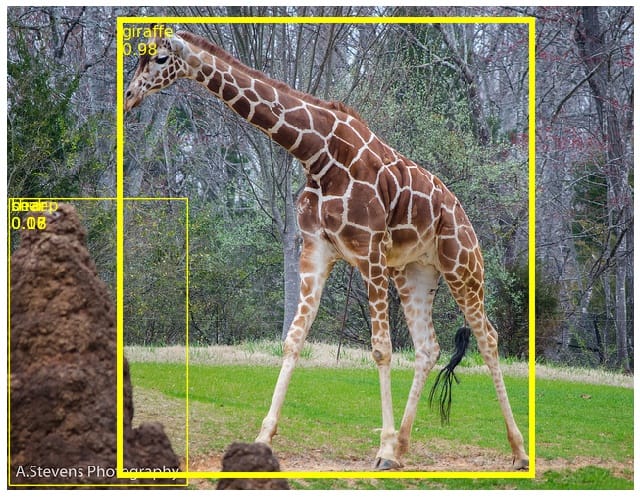}
            \includegraphics[width=0.32\linewidth]{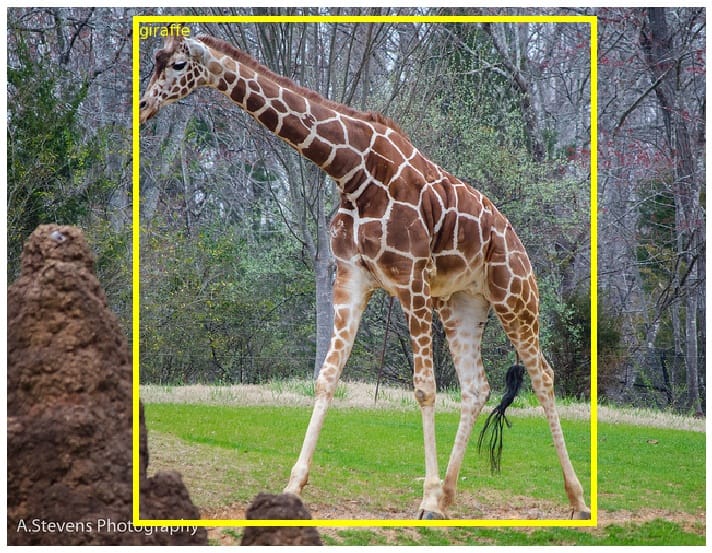}
            \includegraphics[width=0.32\linewidth]{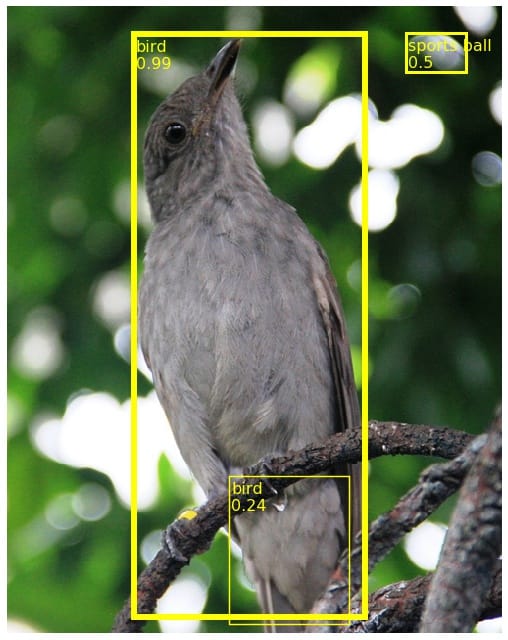}
            \includegraphics[width=0.32\linewidth]{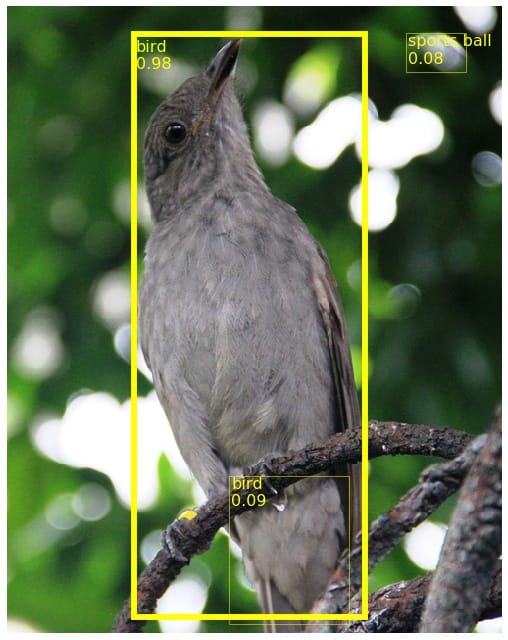}
            \includegraphics[width=0.32\linewidth]{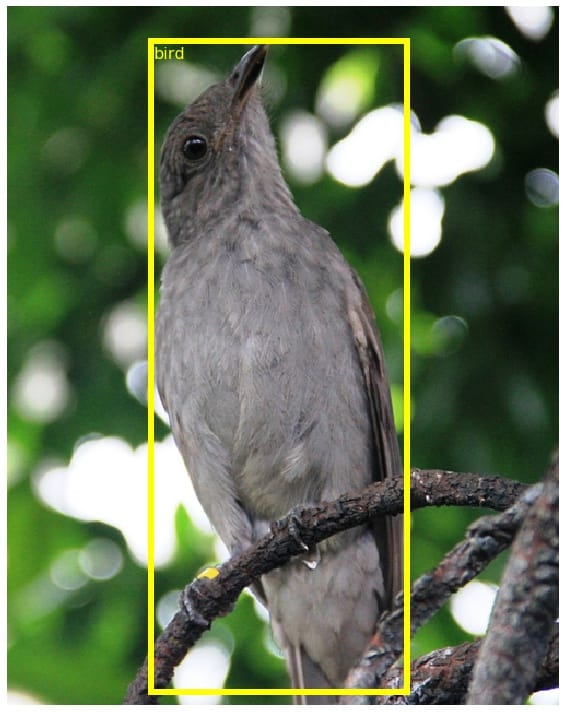}
            \includegraphics[width=0.32\linewidth]{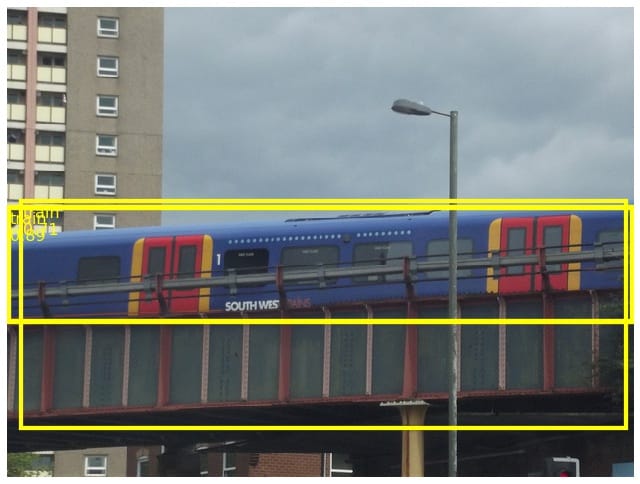}
            \includegraphics[width=0.32\linewidth]{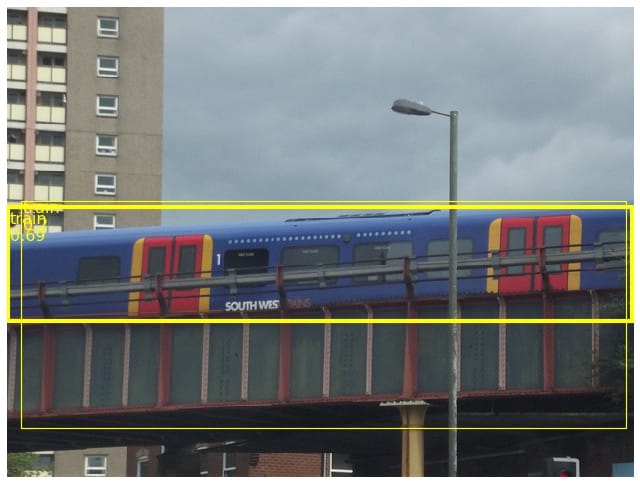}
            \includegraphics[width=0.32\linewidth]{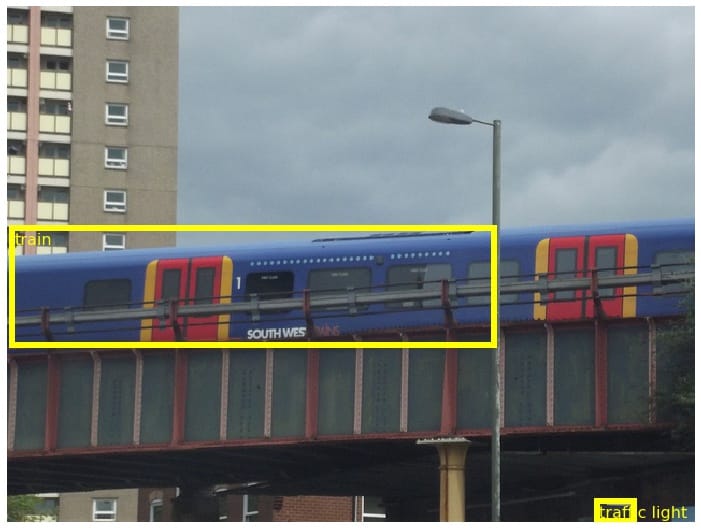}
    \end{minipage}
    \begin{minipage}{.49\linewidth}
            \includegraphics[width=0.32\linewidth]{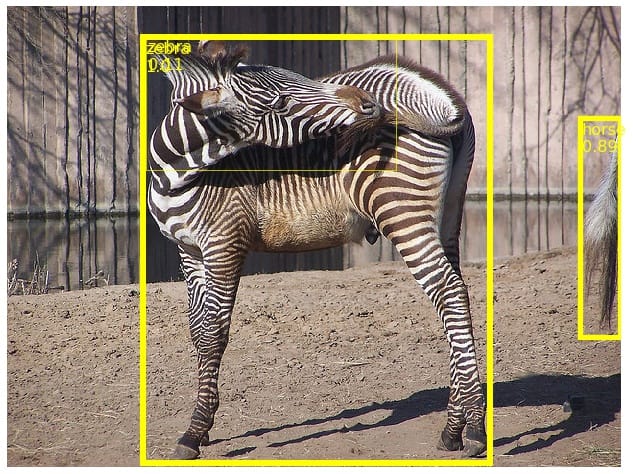}
            \includegraphics[width=0.32\linewidth]{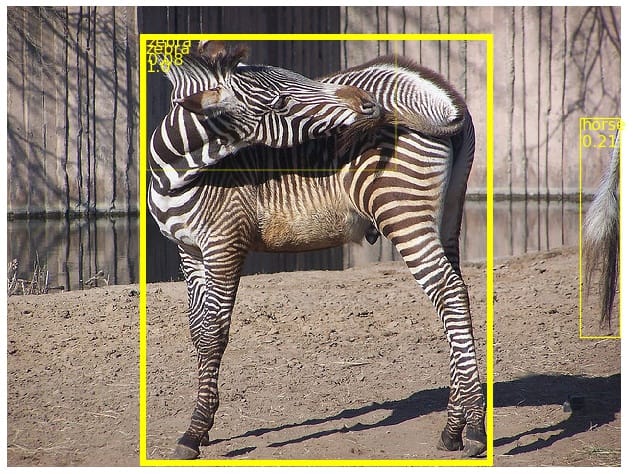}
            \includegraphics[width=0.32\linewidth]{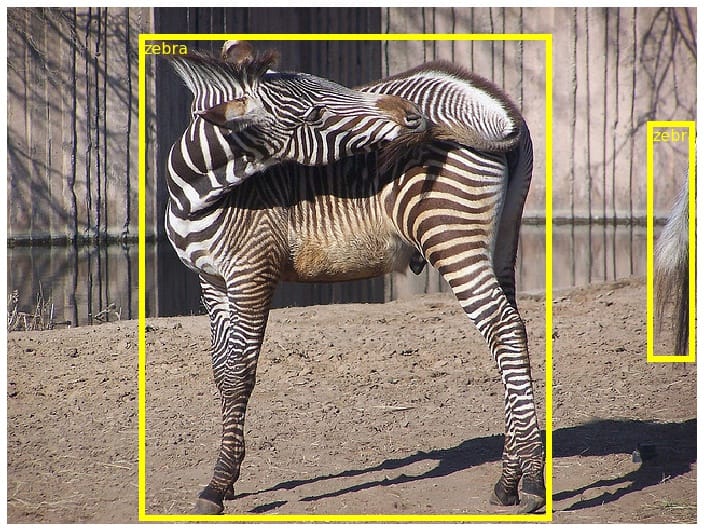}
            \includegraphics[width=0.32\linewidth]{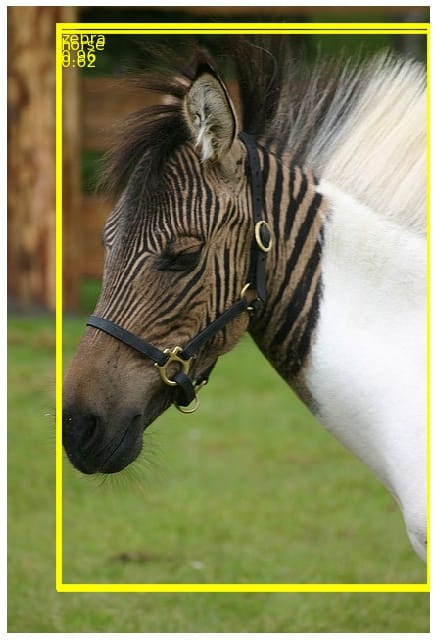}
            \includegraphics[width=0.32\linewidth]{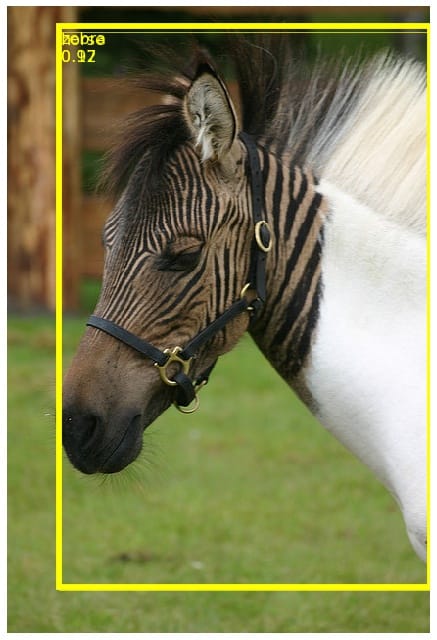}
            \includegraphics[width=0.32\linewidth]{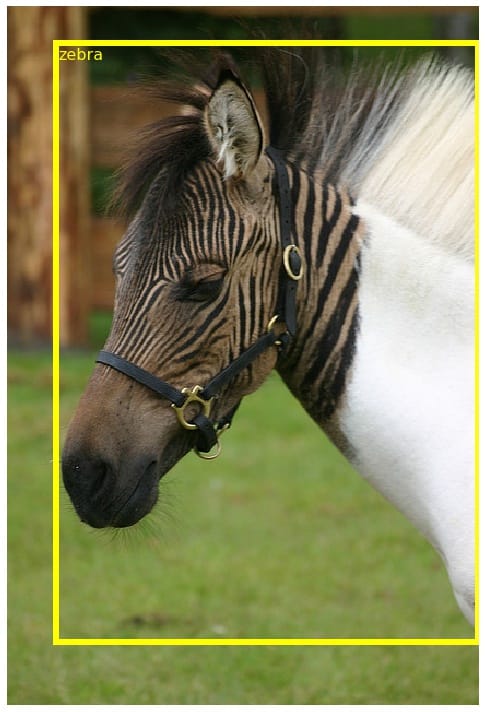}
            \includegraphics[width=0.32\linewidth]{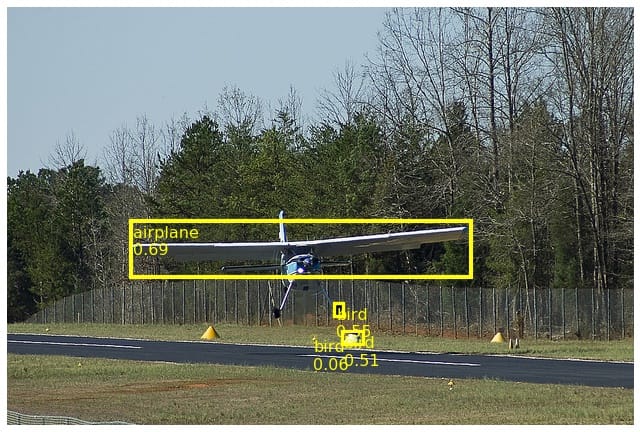}
            \includegraphics[width=0.32\linewidth]{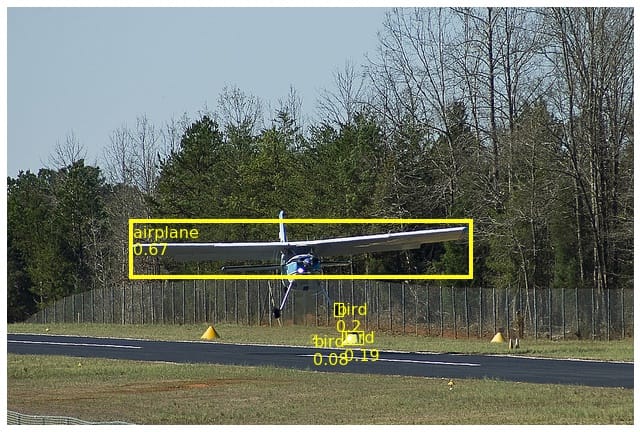}
            \includegraphics[width=0.32\linewidth]{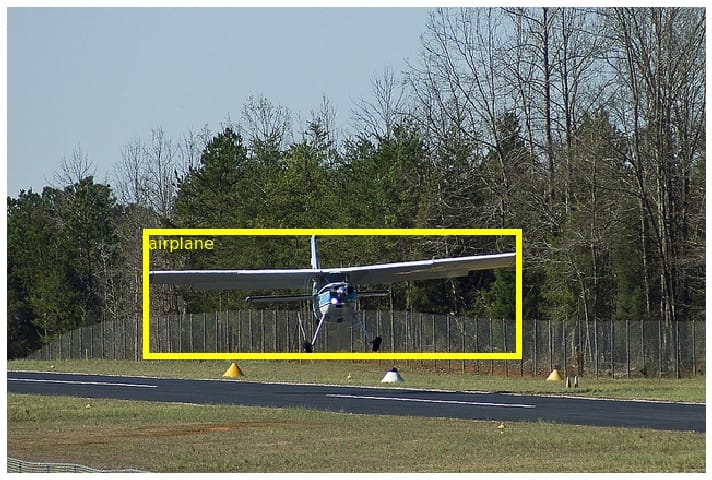}
            \includegraphics[width=0.32\linewidth]{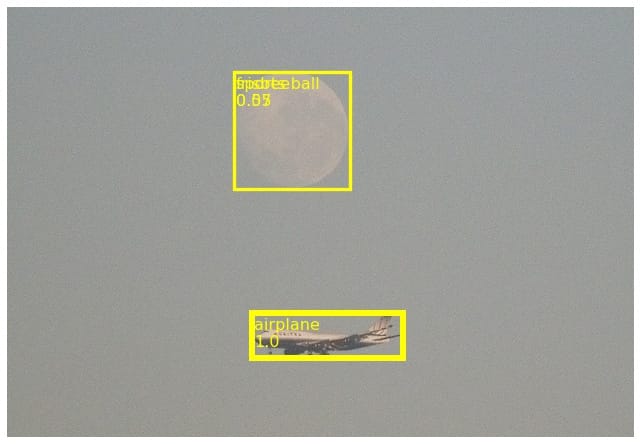}
            \includegraphics[width=0.32\linewidth]{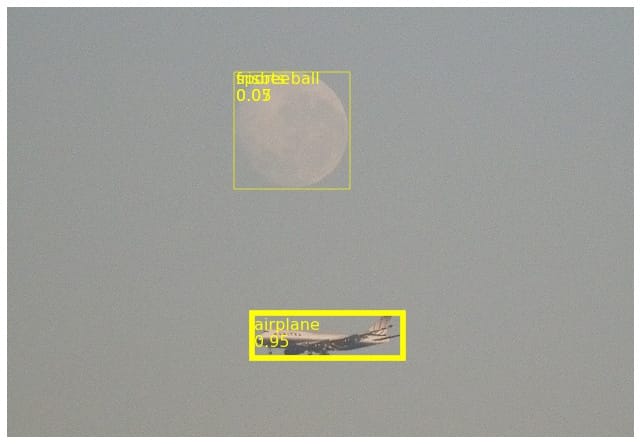}
            \includegraphics[width=0.32\linewidth]{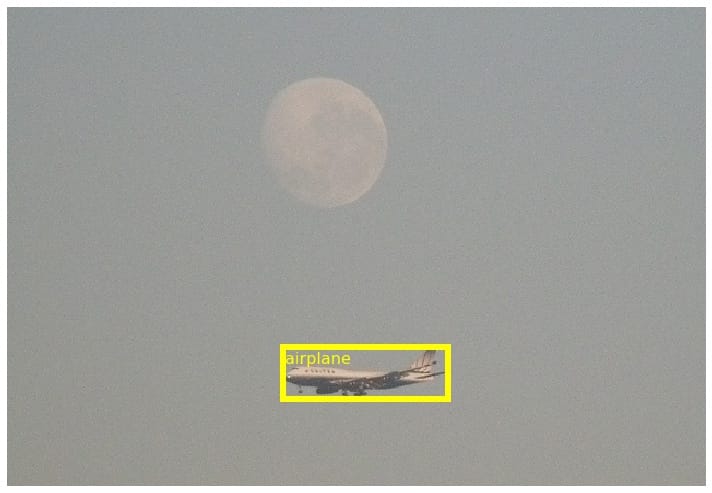}
            \includegraphics[width=0.32\linewidth]{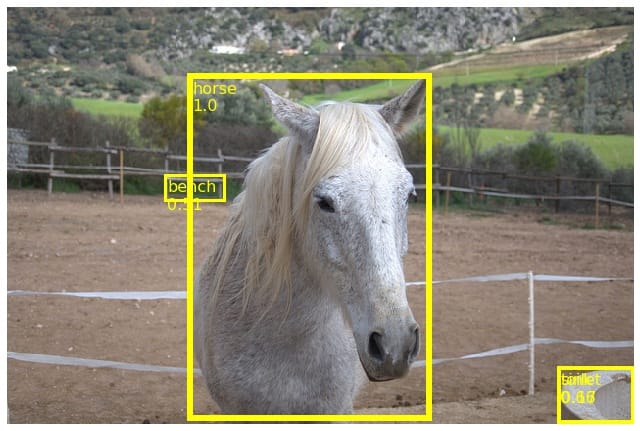}
            \includegraphics[width=0.32\linewidth]{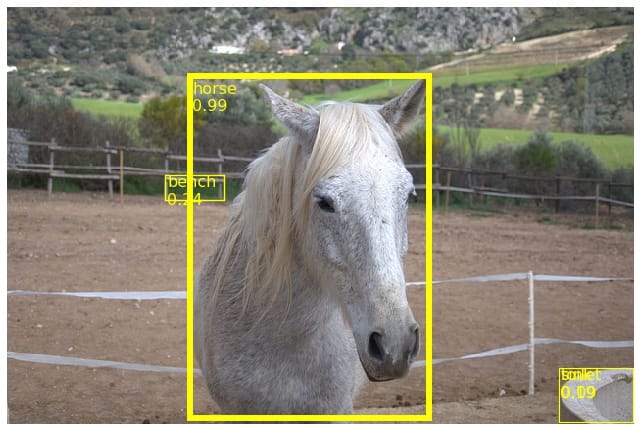}
            \includegraphics[width=0.32\linewidth]{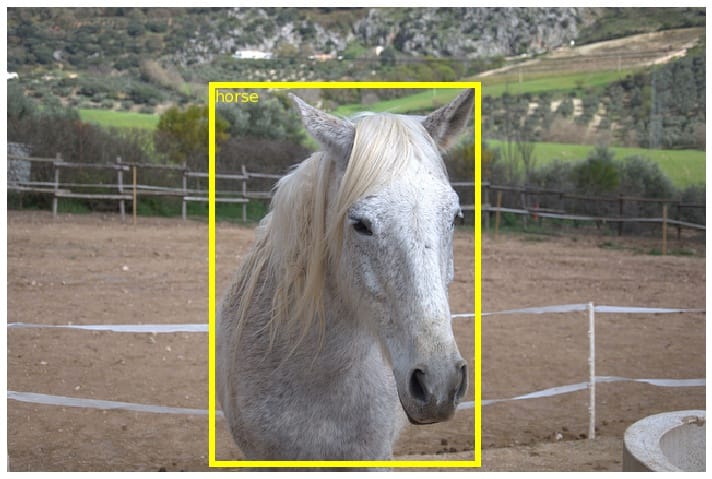}
            \includegraphics[width=0.32\linewidth]{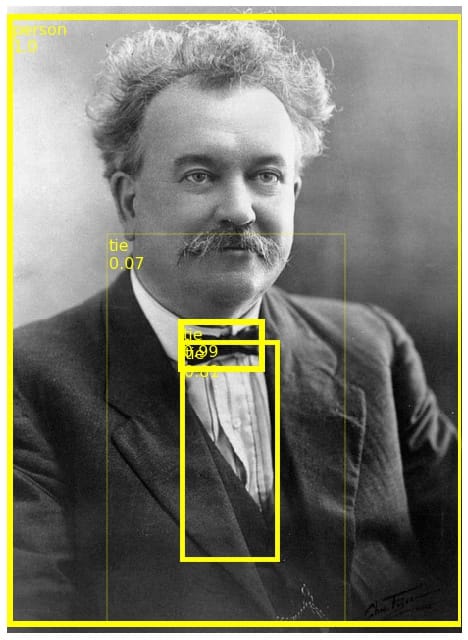}
            \includegraphics[width=0.32\linewidth]{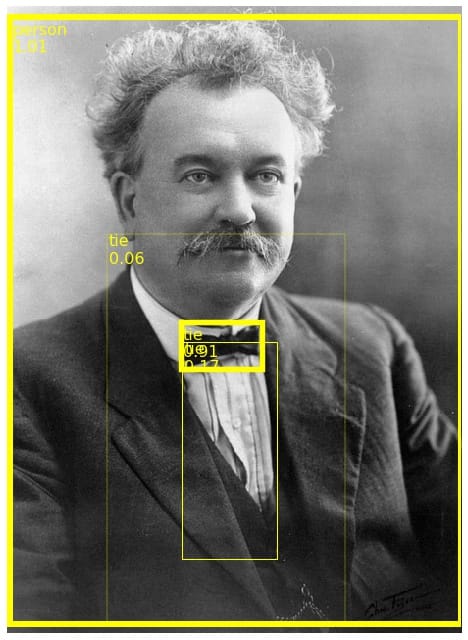}
            \includegraphics[width=0.32\linewidth]{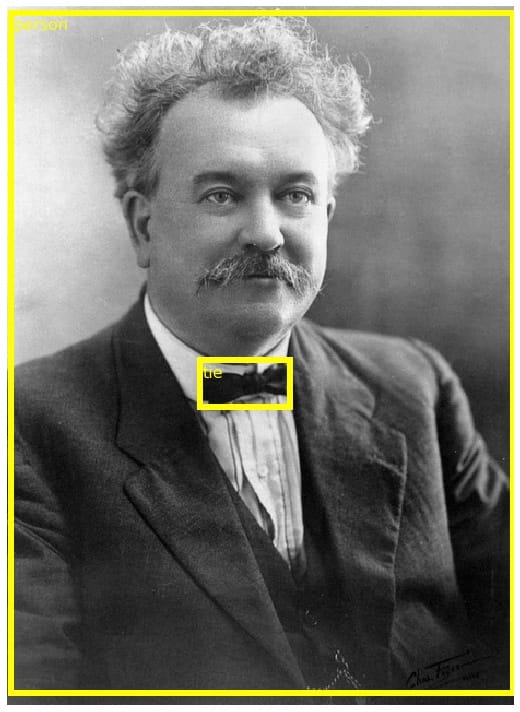}
            \includegraphics[width=0.32\linewidth]{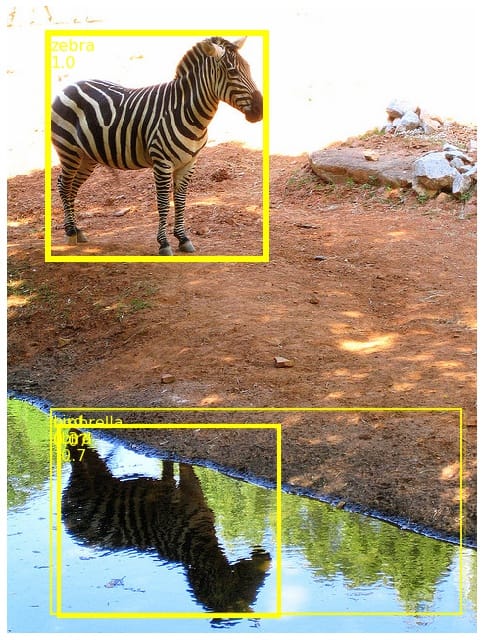}
            \includegraphics[width=0.32\linewidth]{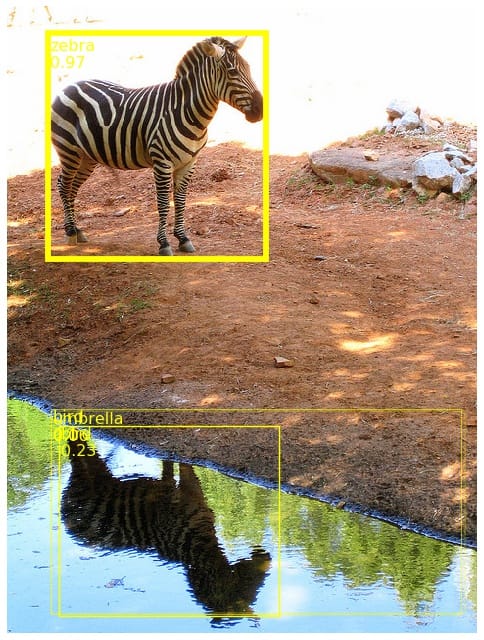}
            \includegraphics[width=0.32\linewidth]{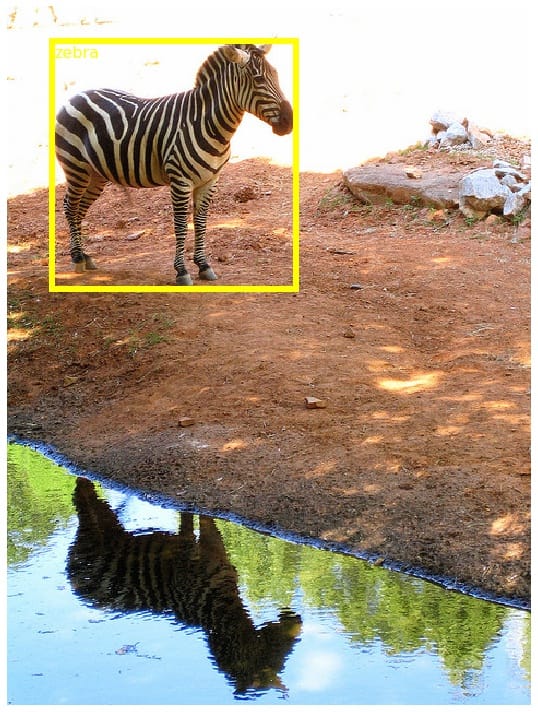}
            \includegraphics[width=0.32\linewidth]{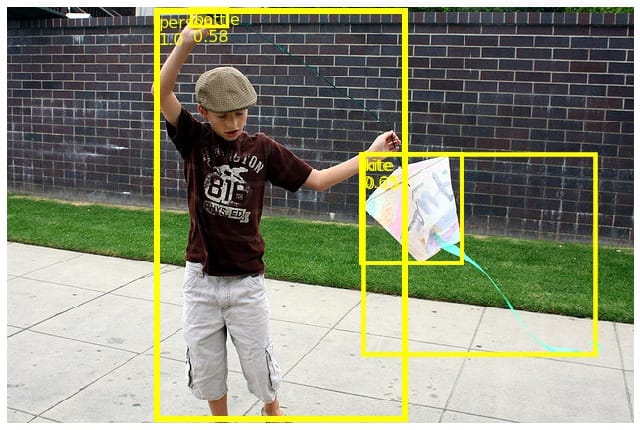}
            \includegraphics[width=0.32\linewidth]{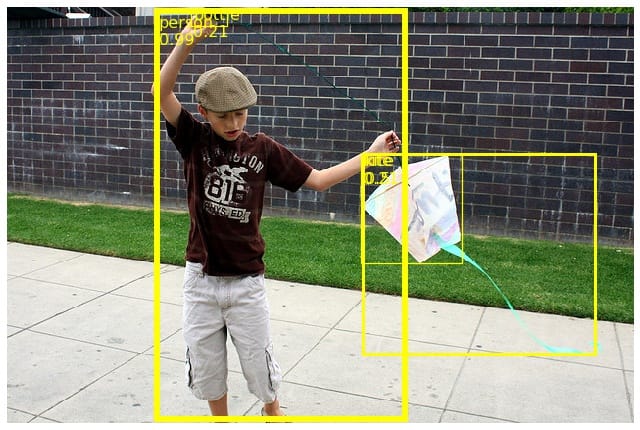}
            \includegraphics[width=0.32\linewidth]{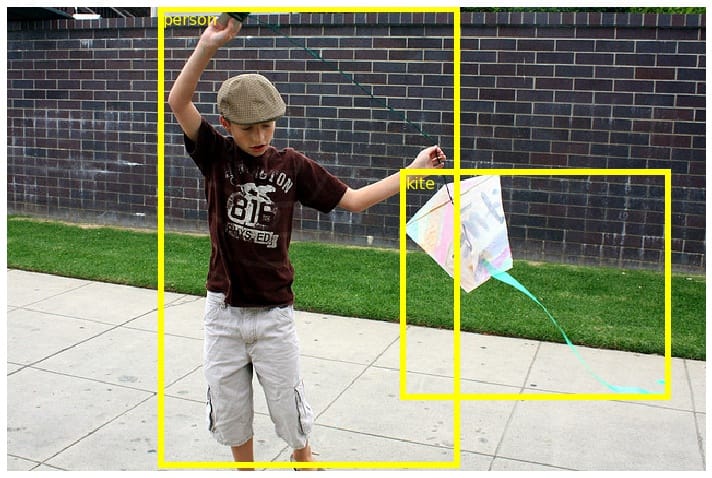}
    \end{minipage}
    \caption{Top $16$ images with at most $4$ detections which had the largest change in confidences as a result of rescoring. For each image, left to right: detections with initial confidences, detections with rescored confidences, and ground truth bounding boxes.}
    \label{fig:top-cosine-max4}
\end{figure*}

\begin{figure*}[tbh]
    \centering
    \begin{minipage}{.49\textwidth}
        \includegraphics[width=0.32\linewidth]{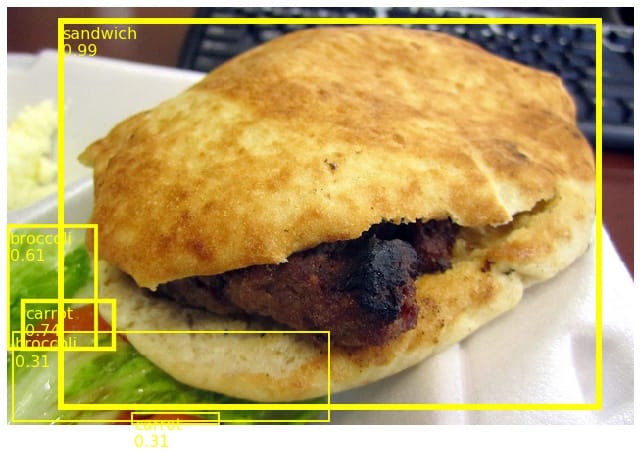}
        \includegraphics[width=0.32\linewidth]{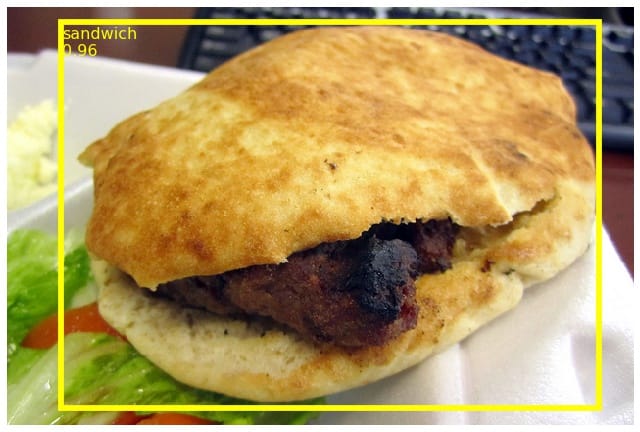}
        \includegraphics[width=0.32\linewidth]{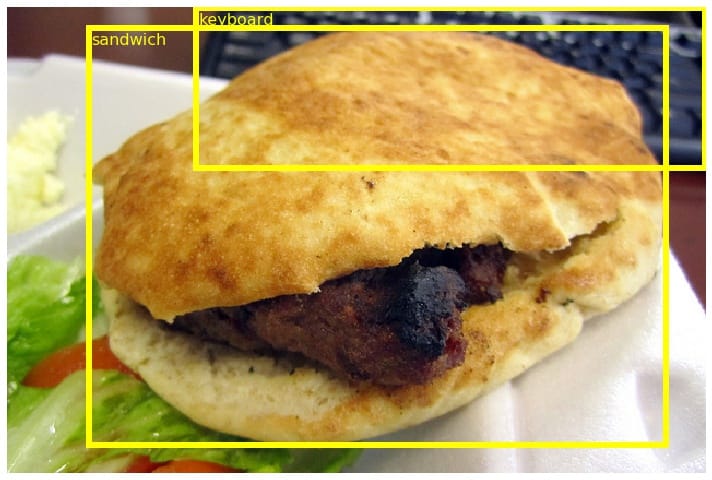}
        \includegraphics[width=0.32\linewidth]{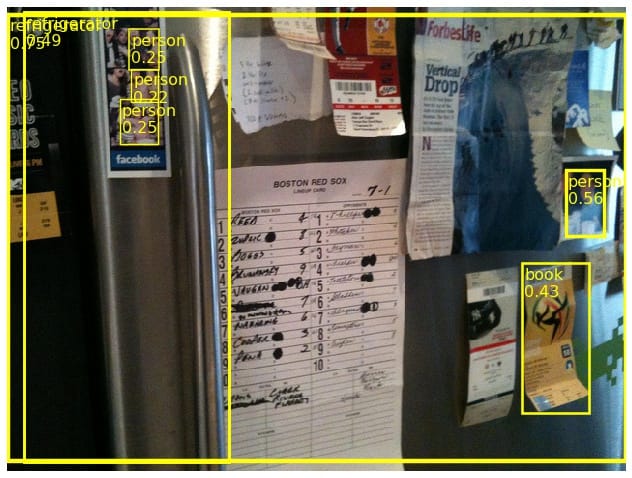}
        \includegraphics[width=0.32\linewidth]{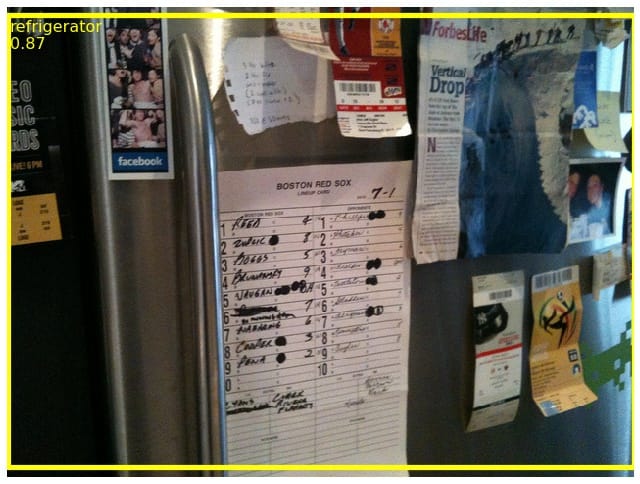}
        \includegraphics[width=0.32\linewidth]{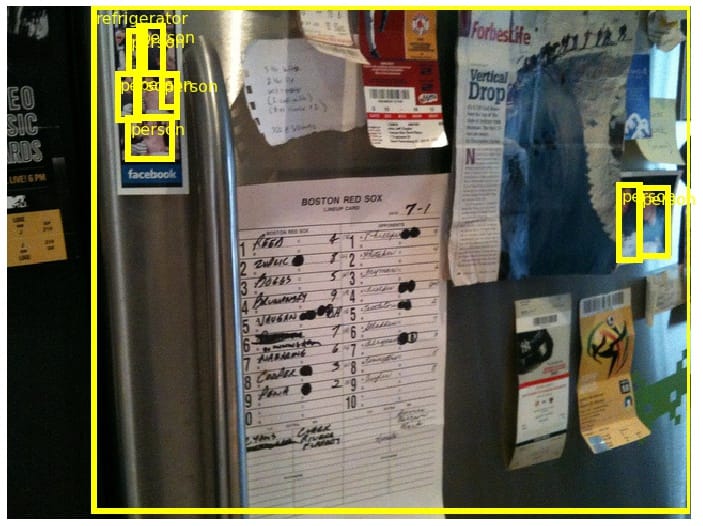}
        \includegraphics[width=0.32\linewidth]{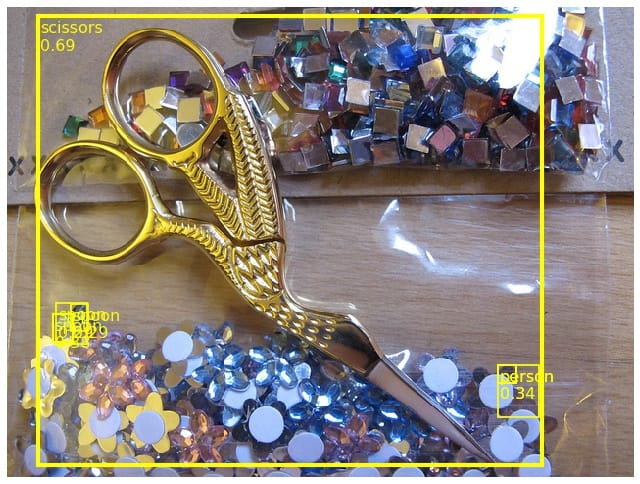}
        \includegraphics[width=0.32\linewidth]{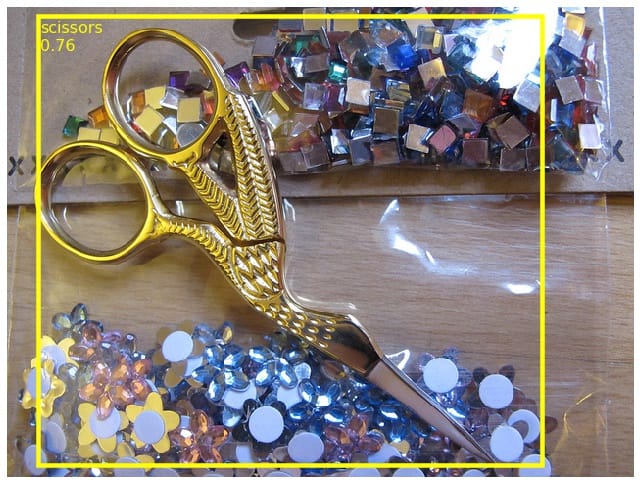}
        \includegraphics[width=0.32\linewidth]{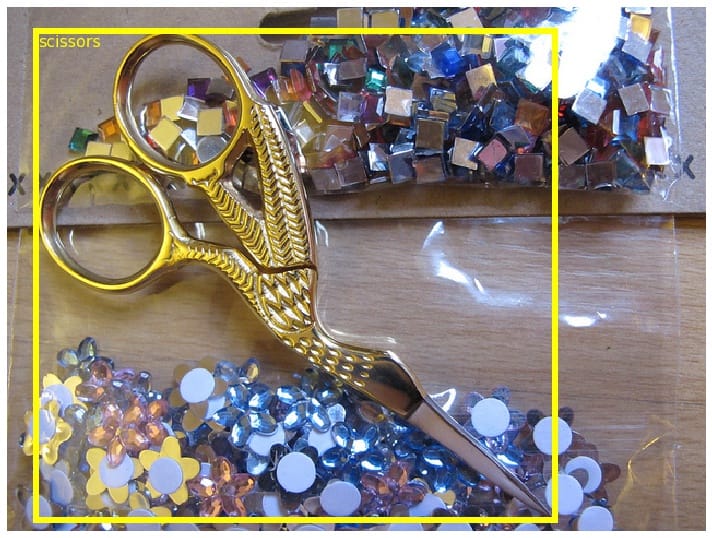}
        \includegraphics[width=0.32\linewidth]{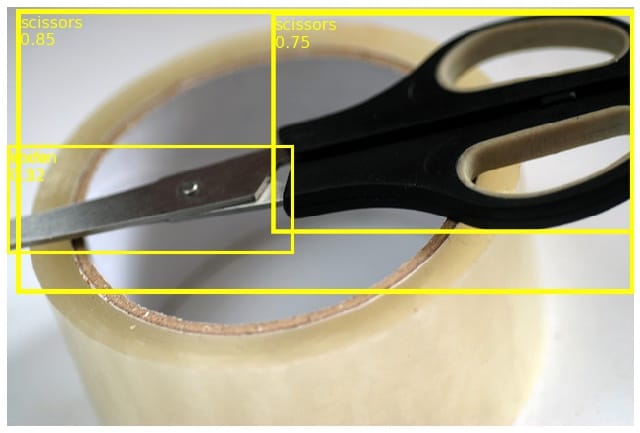}
        \includegraphics[width=0.32\linewidth]{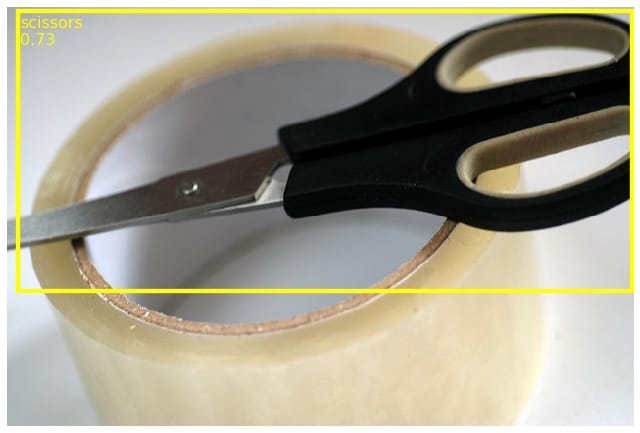}
        \includegraphics[width=0.32\linewidth]{top_cosine_all/6_gold.jpg}
        \includegraphics[width=0.32\linewidth]{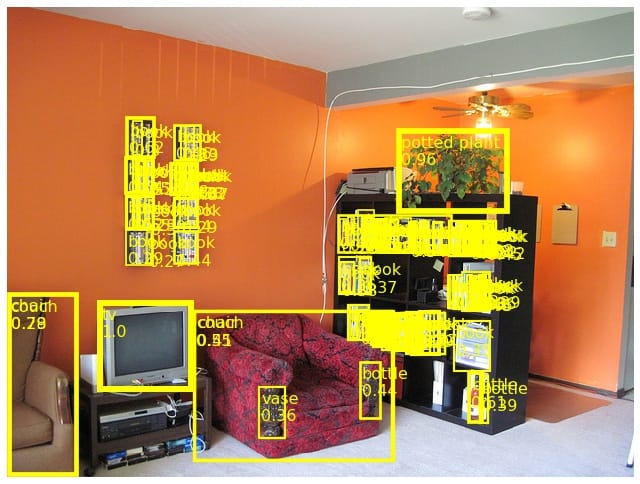}
        \includegraphics[width=0.32\linewidth]{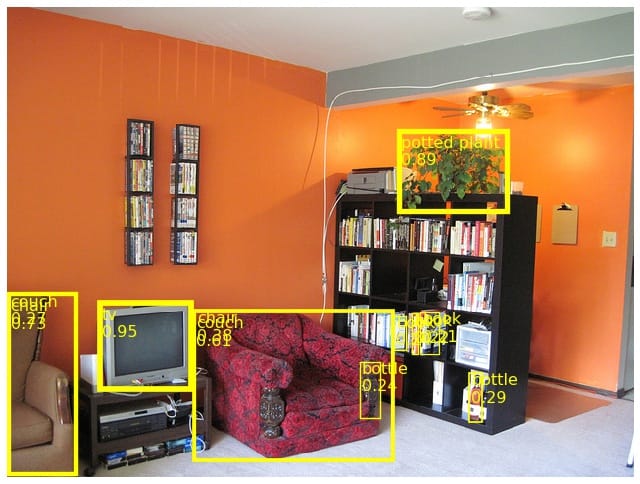}
        \includegraphics[width=0.32\linewidth]{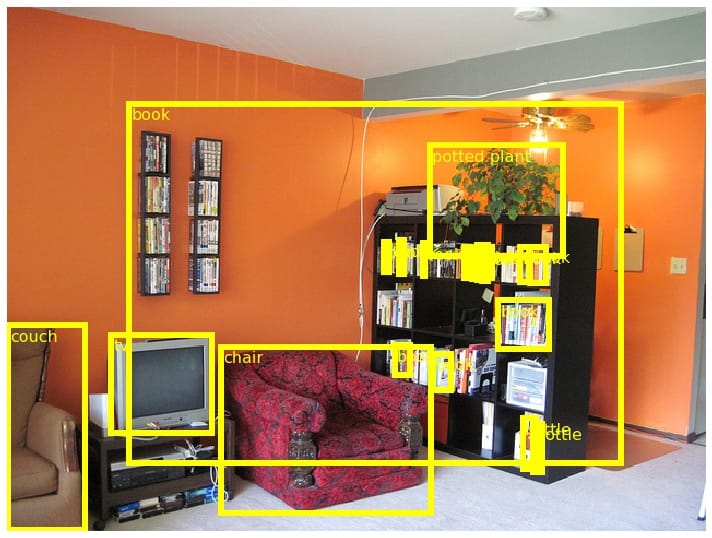}    
        \includegraphics[width=0.32\linewidth]{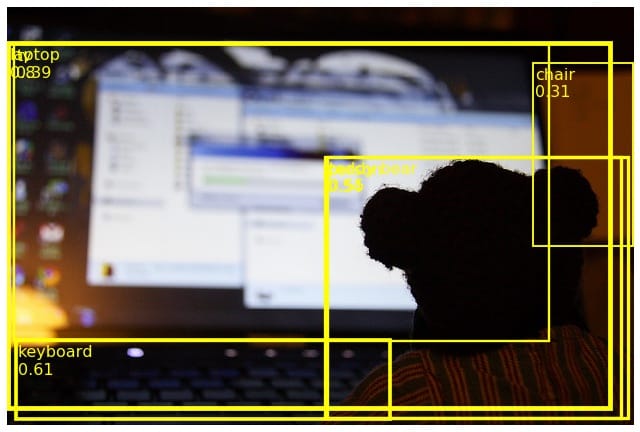}
        \includegraphics[width=0.32\linewidth]{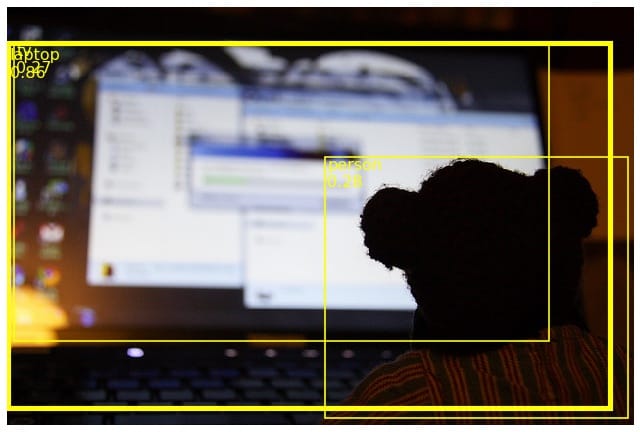}
        \includegraphics[width=0.32\linewidth]{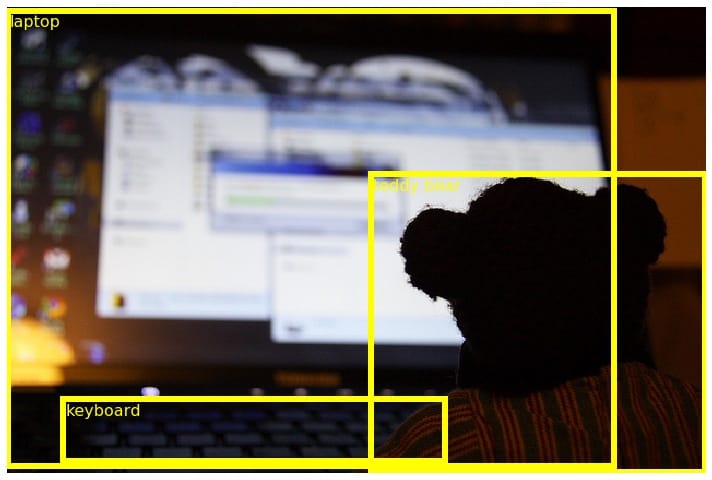}
        \includegraphics[width=0.32\linewidth]{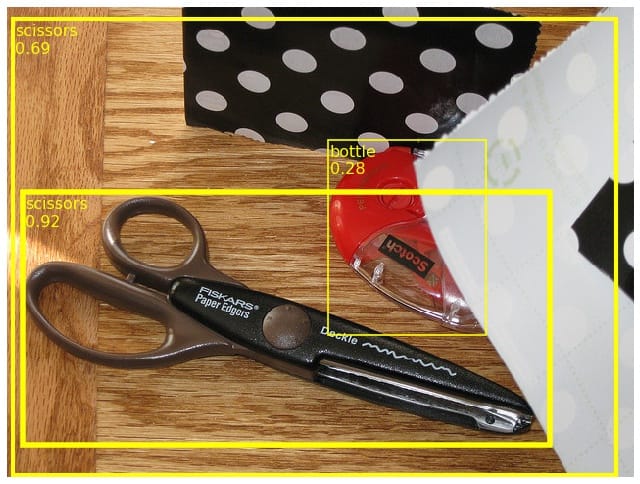}
        \includegraphics[width=0.32\linewidth]{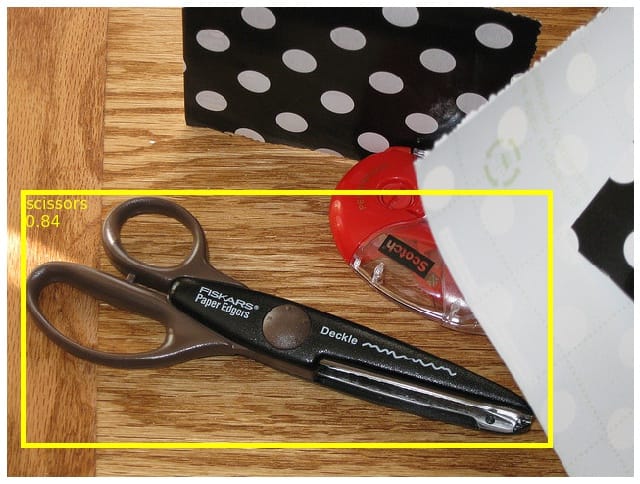}
        \includegraphics[width=0.32\linewidth]{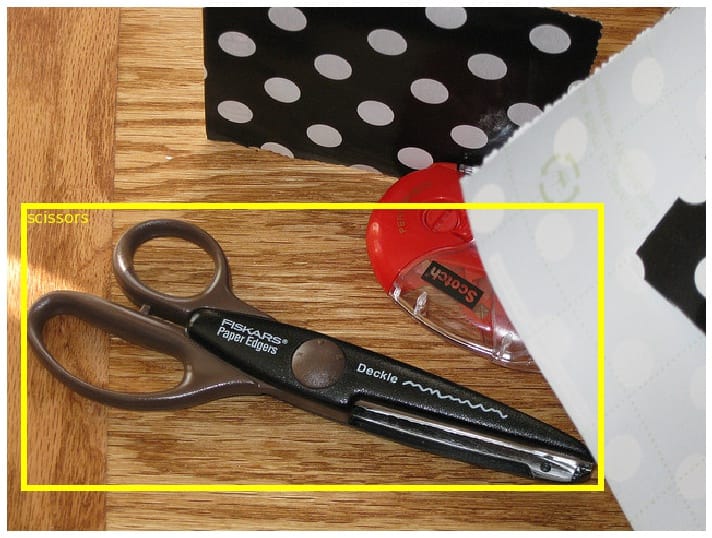}
        \includegraphics[width=0.32\linewidth]{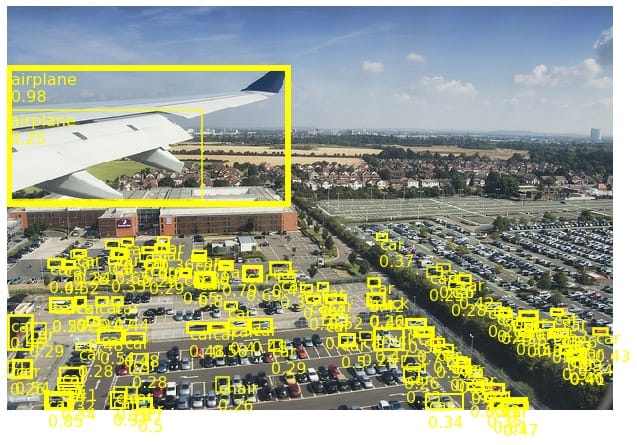}
        \includegraphics[width=0.32\linewidth]{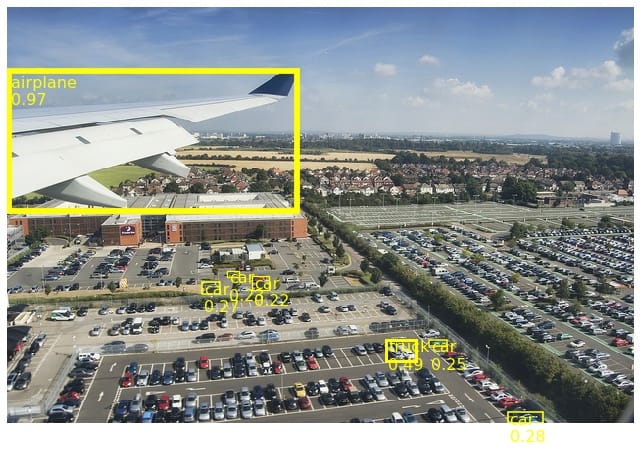}
        \includegraphics[width=0.32\linewidth]{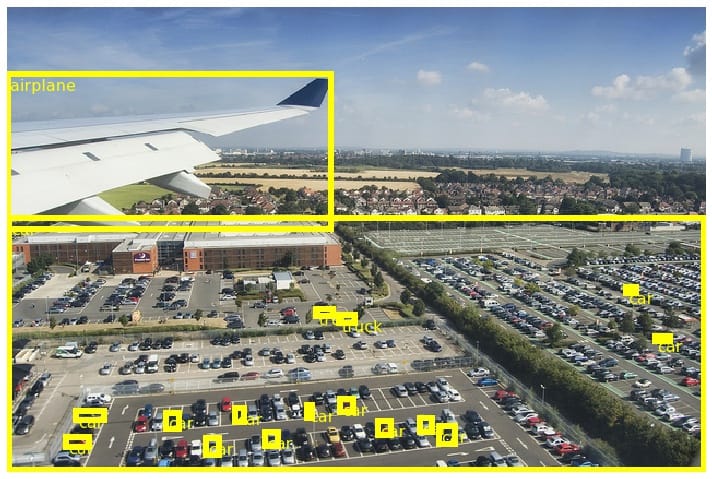}
        \includegraphics[width=0.32\linewidth]{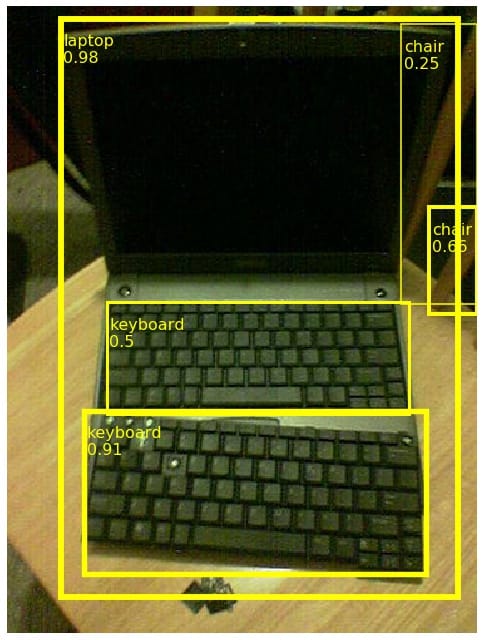}
        \includegraphics[width=0.32\linewidth]{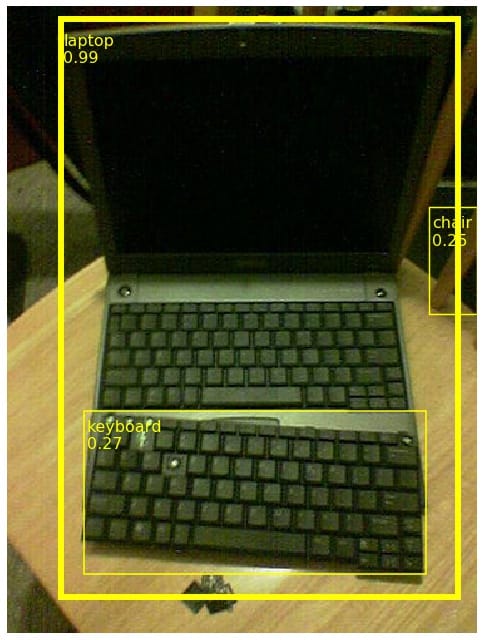}
        \includegraphics[width=0.32\linewidth]{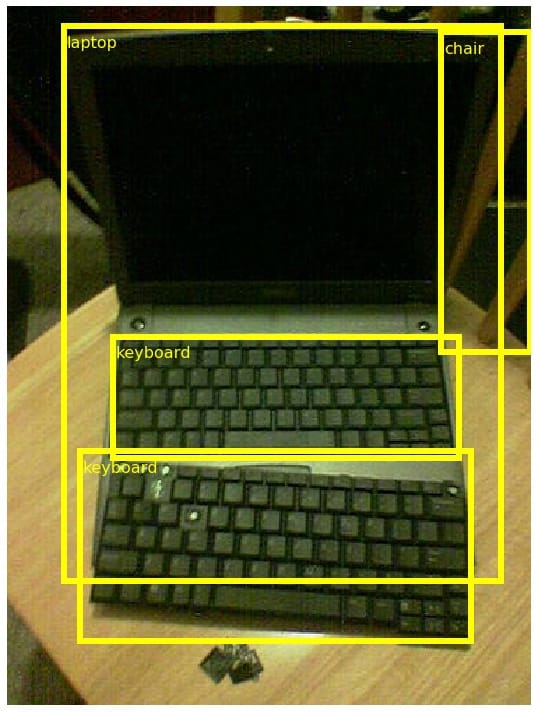}
\end{minipage}
    \begin{minipage}{.49\textwidth}
        \includegraphics[width=0.32\linewidth]{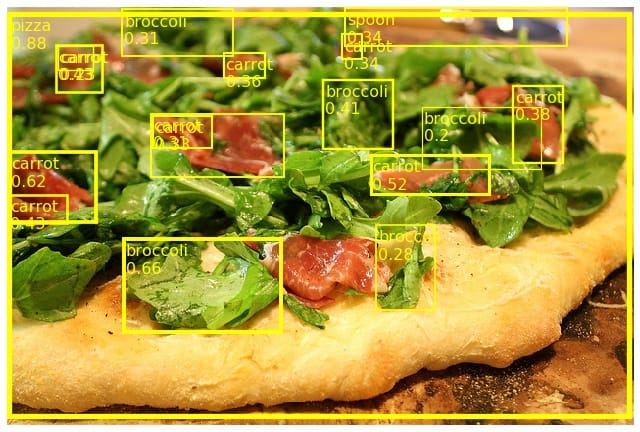}
        \includegraphics[width=0.32\linewidth]{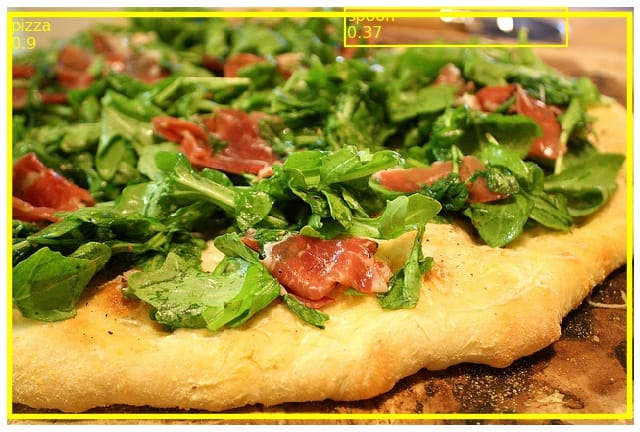}
        \includegraphics[width=0.32\linewidth]{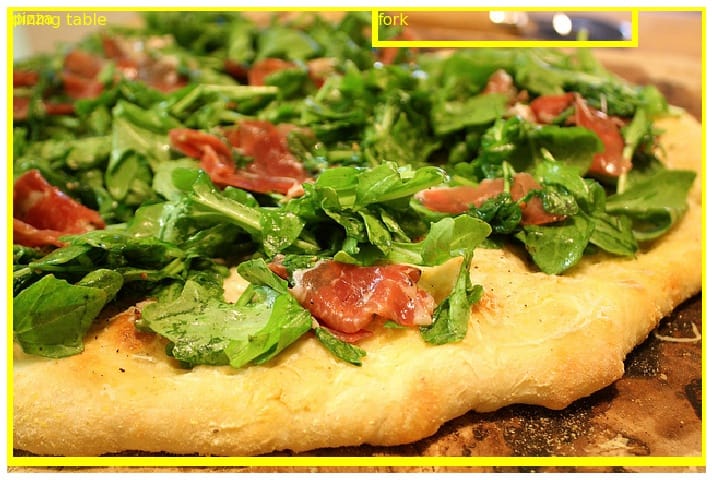}
        \includegraphics[width=0.32\linewidth]{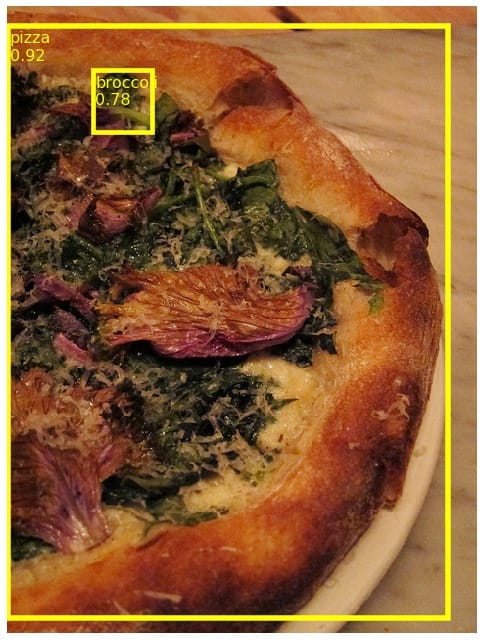}
        \includegraphics[width=0.32\linewidth]{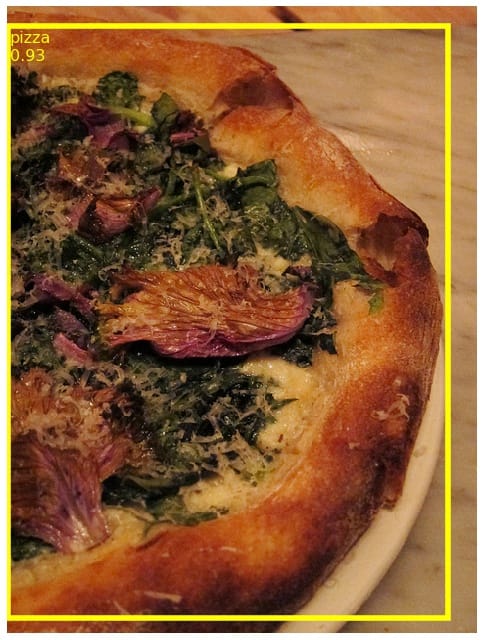}
        \includegraphics[width=0.32\linewidth]{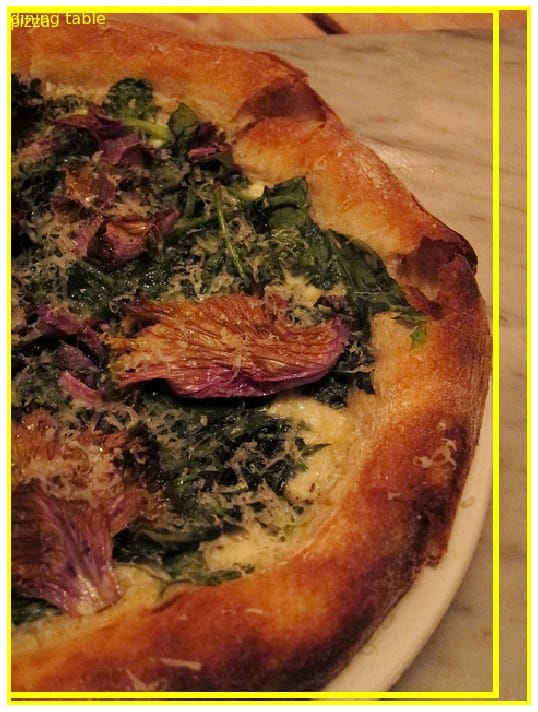}
        \includegraphics[width=0.32\linewidth]{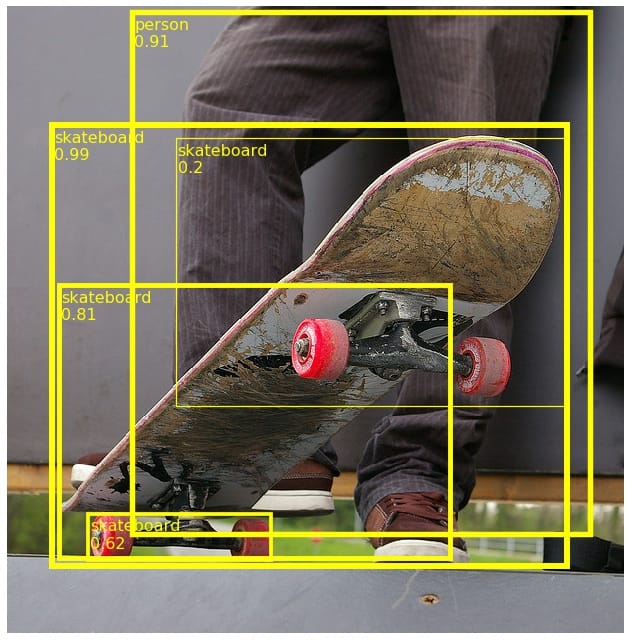}
        \includegraphics[width=0.32\linewidth]{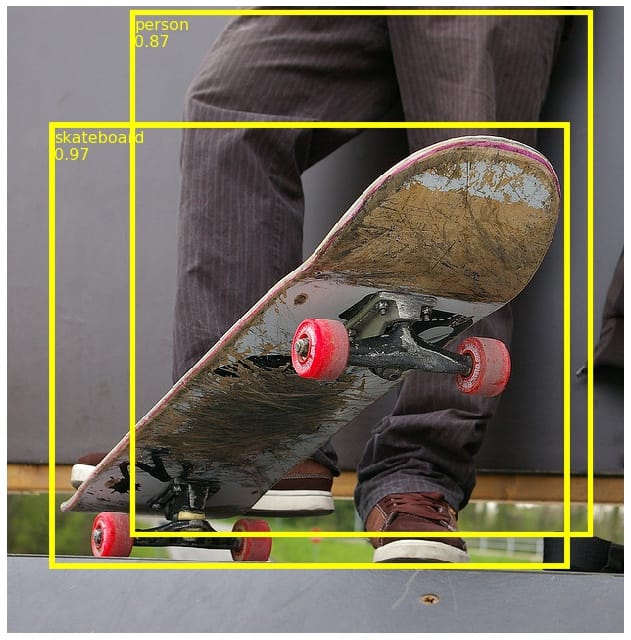}
        \includegraphics[width=0.32\linewidth]{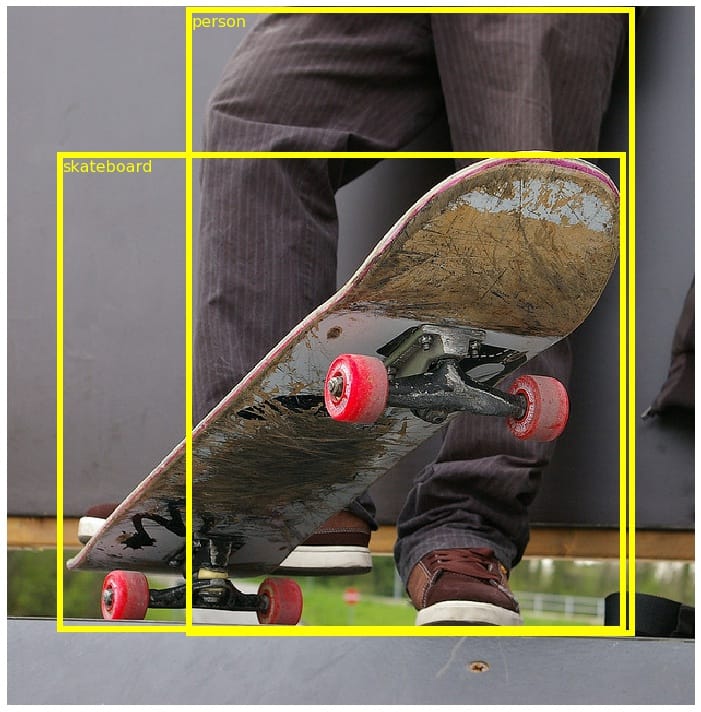}
        \includegraphics[width=0.32\linewidth]{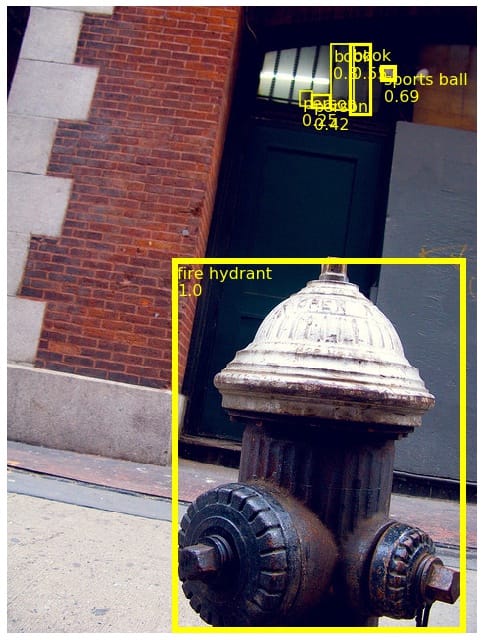}
        \includegraphics[width=0.32\linewidth]{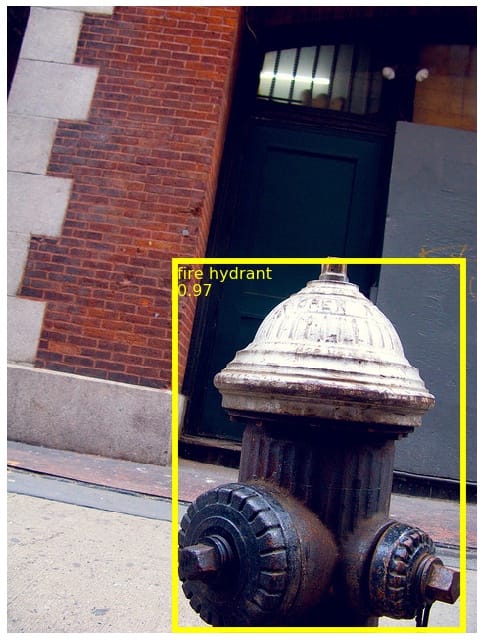}
        \includegraphics[width=0.32\linewidth]{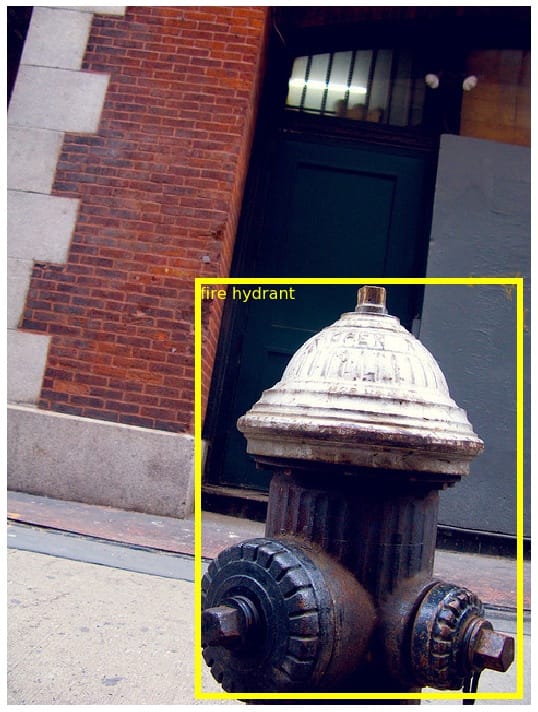}
        \includegraphics[width=0.32\linewidth]{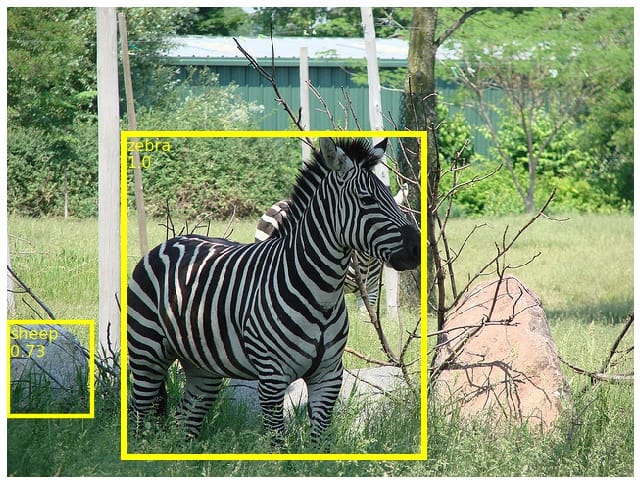}
        \includegraphics[width=0.32\linewidth]{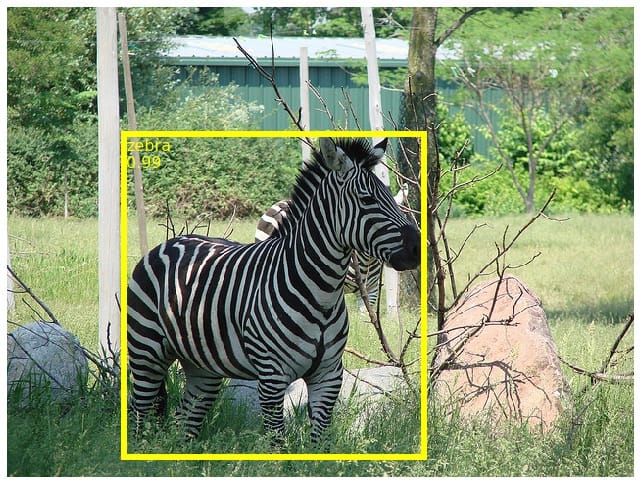}
        \includegraphics[width=0.32\linewidth]{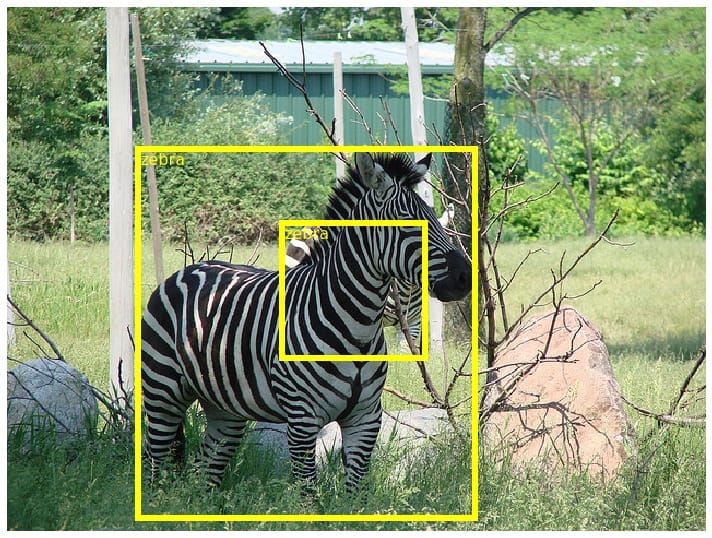}
        \includegraphics[width=0.32\linewidth]{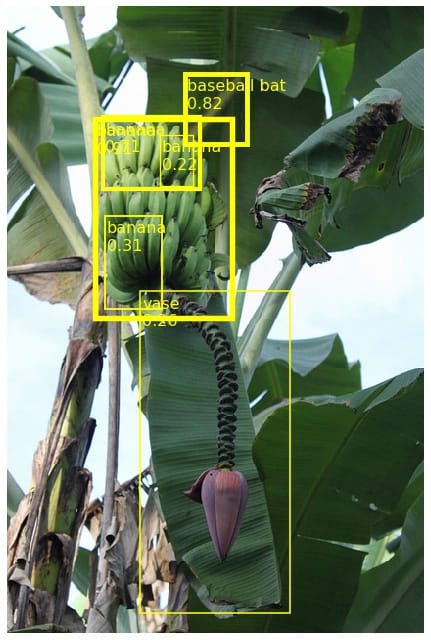}
        \includegraphics[width=0.32\linewidth]{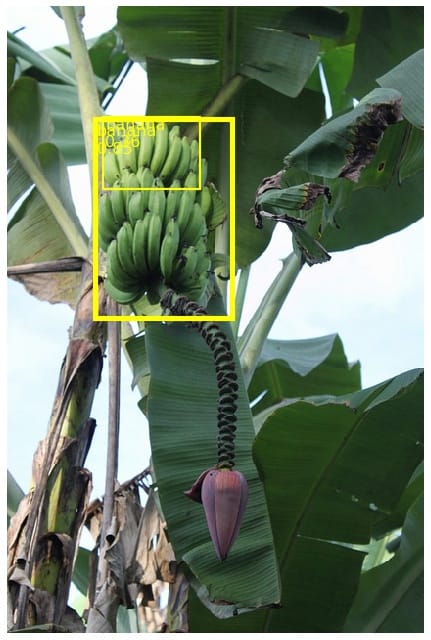}
        \includegraphics[width=0.32\linewidth]{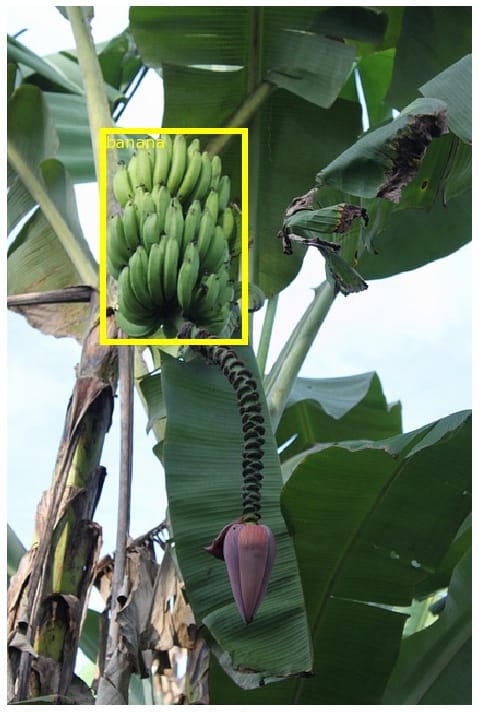}
        \includegraphics[width=0.32\linewidth]{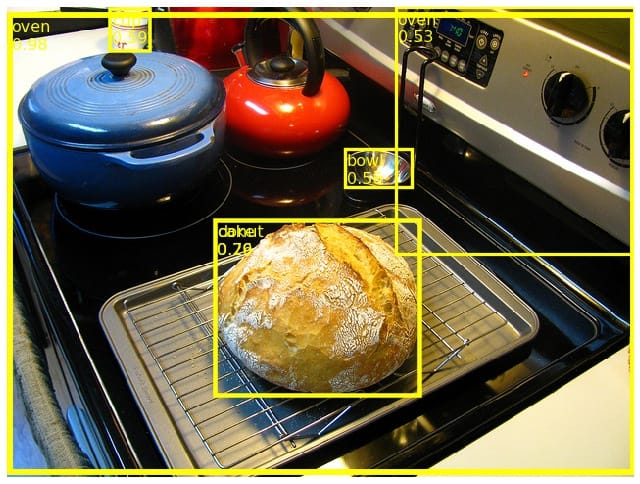}
        \includegraphics[width=0.32\linewidth]{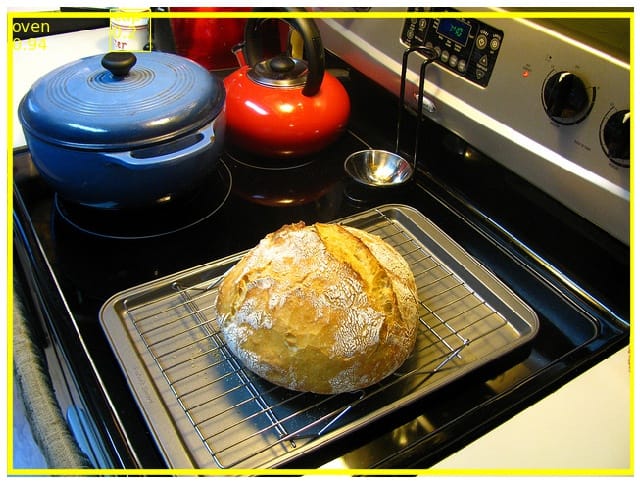}
        \includegraphics[width=0.32\linewidth]{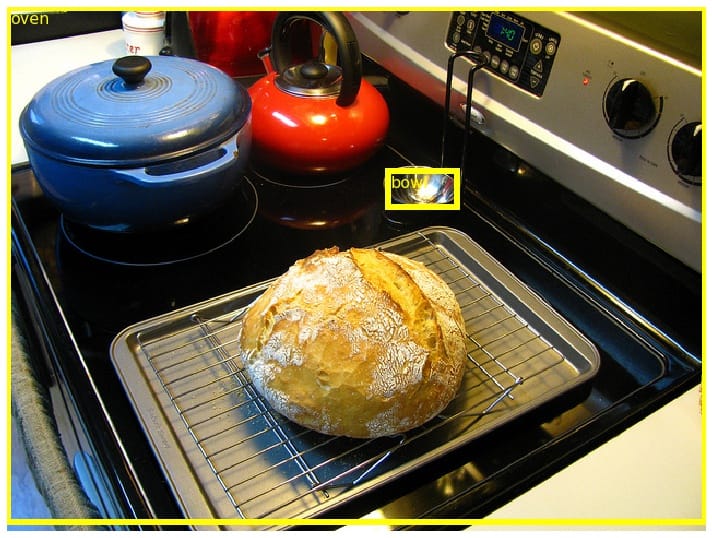}
    \end{minipage}
    \caption{Top $16$ images on which had the largest change in confidences as a result of rescoring. Detections with confidence lower than $0.2$ are omitted. For each image, left to right: detections with initial confidences, detections with rescored confidences, and ground truth bounding boxes.}
    \label{fig:top-cosine-all}
\end{figure*}

We present the top $16$ images according to this metric in two different ways. In Figure~\ref{fig:top-cosine-max4} we only consider images that have at most $4$ detections (i.e., $q \in [n] \mid |\hat{G}_q| \leq 4\}$) as their detections and changes in confidence can be visualized clearly. In Figure~\ref{fig:top-cosine-all}, we consider all images but only show detections that have confidence above $0.2$. An image is shown three times annotated with, left to right, predicted bounding boxes and their confidences before rescoring, predicted bounding boxes and their confidences after rescoring, and ground truth bounding boxes. The bounding box line width is proportional to its confidence. Images are ordered left to right, top to bottom. 

In Figure~\ref{fig:top-cosine-max4}, we see mostly successful suppressions: a rock classified as a sheep in an image with a zebra (left, row 1); duplicate tie detections (left, row 4 and right, row 6); duplicate toilet detections (left, row 2); duplicate train detections (left, row 8); duplicate kite detections (right, row 8); superimposed horse and zebra (right, row 2); duplicate bed detections (left, row 5); the moon classified as a frisbee (right, row 4); a sink and a toilet near a horse (right, row 5); bird and umbrella in the zebra's reflection (right, row 7). 

In Figure~\ref{fig:top-cosine-all}, we have examples with many detections: either for small background objects (left, rows 2, 5 and 8; right, row 4), or multiple duplicate detections of skateboard (right, row 3), banana (right, row 6), and scissors (left, rows 4 and 7). While for most cases we have observed, the model suppresses detections, on the left, on row 3, the model has increased the confidence of its most central object (scissors). In this instance, all original confidences are low (smaller than $0.7$) compared to what happens in most images where there is at least a detection which has more than $0.85$ confidence. 

The behavior of the model shown here can be explained from the point of view of AP computation --- suppressing detections might be useful if we are not confident on their location or existence in the ground truth. This is frequently observed in images containing many (often small) objects of the same class (e.g., apples, bananas, cars, books, and people). The ground truth annotations often do not contain many of the instances in the image. For example, in Figure~\ref{fig:top-cosine-all} (left, row 8), an airplane flies over a parking lot containing hundreds of cars and trucks, out of which only $15$ are in the ground truth annotations. The set of detections contains many of these cars with medium confidence (most ranging from $0.3$ to $0.7$). After rescoring these detections have been mostly suppressed (lower than $0.2$ confidence). 

The reason for this omission in the ground truth annotations is two-fold: perceptually, the exact number of cars is not important and annotating this many cars would be tedious. Due to this, suppressing them during rescoring should lead to improvements as most of these would be considered false positives. The same motivation is valid for the images with books (left, row 5) and bananas (right, row 6). Our approach successfully captures the risk associated with detections being false positives.

\end{document}